%% file: main.tex
\documentclass[11pt]{article}

\usepackage[preprint]{acl}

\usepackage{times}
\usepackage{latexsym}

\usepackage[T1]{fontenc}

\usepackage[utf8]{inputenc}

\usepackage{microtype}

\usepackage{inconsolata}

\usepackage{graphicx}
\usepackage{pgfplots}          
\pgfplotsset{compat=1.18}      
\usepgfplotslibrary{groupplots} 
\usepackage{subcaption}
\usepackage{hyperref}
\usepackage{url}
\usepackage{siunitx}
\usepackage{paralist}
\usepackage{booktabs}
\usepackage{wrapfig}
\usepackage{multirow}
\usepackage{colortbl}
\usepackage{pgfplotstable}
\usepackage[table]{xcolor}
\usepackage{amsmath}
\usepackage{amsfonts}
\usepackage{listings}

\newcommand{\ours}{\textsc{ARM2}}
\newcommand{\method}{\textsc{GRPO-alp}}
\newcommand{\formatOne}{\textit{Direct Answer}}
\newcommand{\formatTwo}{\textit{Short CoT}}
\newcommand{\formatThree}{\textit{Code-Text}}
\newcommand{\formatFour}{\textit{Code-Exec}}
\newcommand{\formatFive}{\textit{Long CoT}}

\title{\ours: Adaptive Reasoning Model with Vision Understanding and Executable Code}

\author{
Jian Xie\textsuperscript{\rm $\spadesuit$}\thanks{Equal contribution.},
Zhendong Chu\textsuperscript{\rm $\heartsuit$}\footnotemark[1],
\textbf{
Aoxiao Zhong\textsuperscript{\rm $\heartsuit$},
Kai Zhang\textsuperscript{\rm $\spadesuit$}},\\
\textbf{
Mingzhe Han\textsuperscript{\rm $\spadesuit$},
Xing Fan\textsuperscript{\rm $\heartsuit$},
Jialie Shen\textsuperscript{\rm $\sigma$},
Qingsong Wen\textsuperscript{\rm $\heartsuit$}}\thanks{Corresponding author.}\\
\textsuperscript{\rm $\heartsuit$}Squirrel Ai Learning 
\textsuperscript{\rm $\spadesuit$}The Ohio State University \\
\textsuperscript{\rm $\sigma$}City St George's, University of London \\
\texttt{\{xie.1741@osu.edu, zc9uy@virginia.edu, qingsongedu@gmail.com\}}
}

\begin{document}
\maketitle
\begin{abstract}
\input{000abstract}
\end{abstract}

\section{Introduction}
\label{sec:intro}
\input{010introduction}

\section{Related Work}
\label{sec:related_work}
\input{020relatedwork}

\section{Methodology}
\label{sec:method}

\input{030method}

\section{Experiments}
\label{sec:experiments}
\input{040experiments}

\section{Results}
\label{sec:results}
\input{050results}

\section{Analysis}
\label{sec:analysis}

\input{060analysis}

\section{Conclusion}
\label{sec:conclusion}
\input{070conclusion}

\bibliography{custom}

\appendix
\clearpage

\section{Appendix}
\label{sec:appendix}
\input{080appendix}

\end{document}

%% file: 000abstract.tex
Large Reasoning Models (LRMs) often suffer from the ``over-thinking'' problem, generating unnecessarily long reasoning on simple tasks. 
Some strategies have been proposed to mitigate this issue, such as length penalties or routing mechanisms, but they are typically heuristic and task-specific, lacking a general framework for adaptive reasoning. 
In this paper, we present \ours{}, a unified model that adaptively balances reasoning performance and efficiency across multiple formats through a reinforcement learning framework augmented with length-aware optimization. 
Beyond conventional natural language inference, \ours{} integrates vision understanding, extending its applicability to multimodal.
Moreover, \ours{} integrates executable code into reasoning, enabling substantial reductions in token cost while preserving task performance compared to long CoT.
Experiments demonstrate that \textbf{\ours{} achieves performance on par with traditional reasoning models trained with GRPO, while reducing token usage by over 70\% on average. }
We further conduct extensive analyses to validate the effectiveness of \ours{} and the soundness of its design.
All the code and resources are released on \href{https://github.com/TEAM-ARM/arm2}{Github}.

%% file: 010introduction.tex
Large Reasoning Models (LRMs)~\citep{jaech2024openai,guo2025deepseek}, trained using Reinforcement Learning with Verifiable Rewards (RLVR), have emerged as a leading paradigm for enhancing the reasoning capabilities of language models.  
Their strong performance is largely attributed to the generous token budget allowed during inference (i.e., long chain-of-thought; long CoT), which enables the use of advanced reasoning strategies such as self-reflection and self-correction~\citep{guo2025deepseek}.  
However, LRMs are not universal ``master keys,'' as they tend to apply long CoT reasoning indiscriminately across tasks, leading to the ``over-thinking'' problem~\citep{sui2025stop,yue2025don}, and in some cases, even causing performance degradation~\citep{wu2024how,fan2025missing}.  
While recent approaches attempt to mitigate the ``over-thinking'' problem, most operate from a single perspective: length control. 
These methods require the model to reason within a fixed token budget or use a separately trained model constrained to specific reasoning lengths~\citep{aggarwal2025l1,alomrani2025reasoning,zhu2025towards}.  
Such strategies assume prior knowledge of the task to predefine the appropriate reasoning length. 
Moreover, they lack the flexibility to switch reasoning formats dynamically across tasks, since manually selecting models or adjusting token limits introduces unnecessary human intervention and hinders generalizability.

In real-world applications, models are often required to balance both low-latency responses and complex reasoning. 
For example, for relatively simple questions, models are expected to provide direct answers without unnecessary deliberation. 
In contrast, for more challenging questions, models are expected to generate reliable answers through comprehensive reasoning.
This raises a critical research question: \emph{How can we equip LRMs with the ability to perform adaptive reasoning based on task demands and input complexity?}

Hybrid reasoning strategies~\citep{lou2025adacot,zhang2025adaptthink,wu2025arm} provide viable approaches to balance performance and efficiency.
For example, ARM~\citep{wu2025arm} introduces four distinct reasoning formats and encourages the use of more efficient formats rather than relying exclusively on long CoT.  
However, these approaches primarily focus on incorporating shorter reasoning formats to mitigate the substantial token cost of long CoT, without explicitly leveraging length-aware reward signals, which limits their effectiveness in reducing token usage.
Furthermore, existing adaptive reasoning models are designed for plain-text inputs and lack the ability to handle visual information.
In parallel, recent advances in coding capabilities and their integration into reasoning~\citep{chen2022program,wang2024executable} have highlighted code execution during inference as a powerful mechanism for enhancing reasoning. 
Incorporating executable code enables models to perform verifiable computations, manage complex logic, and interact with external tools, thereby extending their reasoning capacity well beyond traditional text-based methods.

In this paper, as shown in Figure~\ref{fig: arvec}, we propose \ours{}---an adaptive reasoning model with vision understanding and executable code. 
\ours{} supports five reasoning formats: \formatOne{}, \formatTwo{}, \formatThree{}, \formatFour{}, and \formatFive{}. 
Given a task, \ours{} adaptively selects one of these formats for reasoning. 
Among them, \formatOne{}, \formatTwo{}, \formatThree{}, and \formatFive{} are based on natural language inference, while \formatFour{} leverages an external Python interpreter to execute code and return the result as the final answer. 
This capability empowers the model to solve complex reasoning tasks that require high precision, thereby extending its reach beyond conventional natural language inference.

To enable adaptive reasoning, we curate a high-quality multimodal dataset of \num{15.1}K instances spanning all five reasoning formats, providing a strong foundation for cold-start through supervised finetuning (SFT). 
However, since SFT only teaches the use of different reasoning formats without enabling adaptive selection, we turn to reinforcement learning (RL). 
Yet conventional Group Relative Policy Optimization (GRPO)~\citep{shao2024deepseekmath} suffers from format collapse~\citep{wu2025arm}, where models tend to over-rely on the reasoning format with the highest accuracy.
To overcome this issue, we propose \method{}, a novel variant of GRPO designed to promote adaptive reasoning by encouraging exploration of less frequent formats with relatively lower accuracy during training. 
While related to Ada-GRPO~\citep{wu2025arm}, which also addresses format imbalance, our approach goes further by explicitly incorporating response length into the optimization objective. 
This enables \method{} not only to achieve balanced format selection but also to guide the model in trading off reasoning accuracy against computational efficiency.
In comparison to traditional LRMs, \ours{} demonstrates improved efficiency on straightforward, intuitive tasks while preserving competitive performance on reasoning-intensive tasks.

In summary, our contributions are as follows:
\begin{inparaenum}[\it 1)]
\item \ours{} achieves a strong balance between performance and efficiency, reducing token usage by over 70\% on average across six in-domain and six out-of-domain datasets, spanning both text and multimodal settings.
\item \ours{} first incorporates executable code into adaptive reasoning, improving code reasoning accuracy and providing a more efficient alternative to long CoT reasoning.
\item We conduct systematic experiments to validate the design of \ours{}, including ablation studies on its components as well as analyses of length penalty strength and backbone model initialization.
\end{inparaenum}

%% file: 020relatedwork.tex
\subsection{Overthinking}
Capable of performing elaborative reasoning at inference time, Large Reasoning Models (LRMs) achieve substantial performance gains across a wide range of reasoning benchmarks~\citep{jaech2024openai,guo2025deepseek}. 
However, not all tasks require such an intensive reasoning style, and prior studies have shown that LRMs often incur unnecessary token overhead on relatively simple tasks~\citep{chen2024not}. 
More critically, this inefficiency not only leads to computational waste but also undermines the model's overall effectiveness, as it may produce misleading or confusing content, such as repeated reflections, unwarranted self-doubt, or overconfident simulation~\citep{fan2025missing,cuadron2025danger}, which could have been avoided with more direct responses~\citep{alomrani2025reasoning}.
This challenge is particularly troublesome because the very strength of LRMs lies in their extended reasoning steps, making their resolution appear paradoxical~\citep{sui2025stop}.

\begin{figure*}[t]
\centering
    \includegraphics[width=\linewidth]{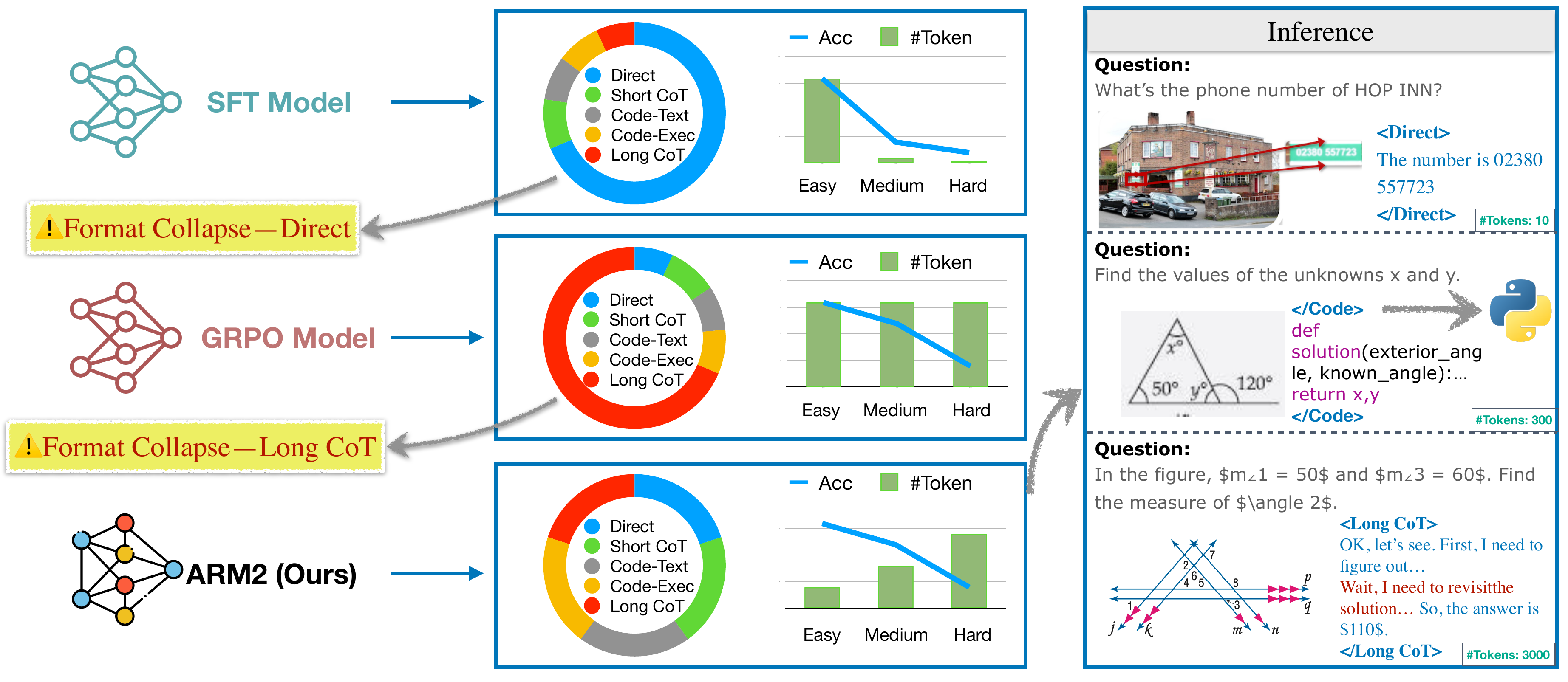}
    \caption{Comparison of reasoning behaviors across models. Unlike the SFT and GRPO models, which consistently rely on a single reasoning format, \ours{} exhibits adaptability by selecting reasoning formats.}
    \label{fig: arvec}
\end{figure*}

\subsection{Controllable and Adaptive Reasoning}
To improve the efficiency of LRMs, previous work has incorporated length constraints into the reward function~\citep{aggarwal2025l1,luo2025o1,hou2025thinkprune} under reinforcement learning optimization, encouraging shorter responses while maintaining correctness. 
While this strategy can indeed reduce response length at inference time, it has a key limitation: it relies on accurately estimating the task's complexity. 
If the estimation is imprecise, the model's performance may be significantly affected~\citep{wu2025arm}.
Another line of research focuses on adaptive reasoning, which allows models to switch among different reasoning formats, such as short CoT for easy tasks and long CoT for more complex ones~\citep{zhang2025adaptthink,luo2025ada}. 
However, this strategy also faces two major limitations: the restricted set of reasoning styles available for handling diverse tasks, and the reliance on task-specific supervision during training without incorporating length awareness into reasoning control.
To address these challenges, we propose \ours{}, a unified model capable of flexibly switching among five reasoning styles for different tasks, while explicitly incorporating length awareness into the reasoning process.

%% file: 030method.tex
In this paper, we introduce \ours{}, a multi-modal reasoning model that adaptively adjusts its reasoning formats based on tasks while maintaining consistent performance compared to vanilla reasoning models.
To train \ours{}, we adopt the following training pipeline:
\begin{inparaenum}[\it 1)]
\item
Data Construction, building multi-modal QA datasets with diverse reasoning formats;
\item
Supervised Fine-tuning (SFT), teaching the model to reason using different formats;
\item
Length-aware Reinforcement Learning (RL), enabling the model to select the most suitable reasoning format to balance  performance and token efficiency; 
and \item Inference, applying the trained model for adaptive reasoning across tasks.
\end{inparaenum}

\subsection{Data Construction}
Our training dataset consists of both a plain-text reasoning part and a multi-modal reasoning part. 
For the plain-text part, we adopt the AQuA-Rat dataset~\citep{ling2017program} used in ARM~\citep{wu2025arm}.
To activate the multi-modal reasoning capability of \ours, we further collect data from VisualWebInstruct~\citep{jia2025visualwebinstruct} to extend the modality to vision.  
While \formatOne{} and \formatTwo{} rationales are directly provided within the dataset, we augment it with \formatThree{}, \formatFour{}, and \formatFive{} rationales using GPT-4o~\citep{gpt4o} and Doubao-Pro-1.6-Thinking~\citep{seed2025seed1}.  
Moreover, all rationales in the code reasoning format are verified for executability using an external interpreter.  
This enables \ours{} to support executable code generation during inference.  
To ensure high-quality supervision, we filter out instances whose rationales yield incorrect answers, resulting in a curated dataset of 15.1K distinctive training examples spanning all reasoning formats.

For the reinforcement learning stage, the flexibility of RLVR allows us to leverage datasets that contain only questions and answers.  
Specifically, we incorporate three text-only datasets: CommonSenseQA (CSQA)~\citep{talmor2019commonsenseqa}, GSM8K~\citep{cobbe2021training}, and AIME 1987-2023~\citep{aime25}, covering a spectrum from simple commonsense reasoning to complex mathematical problem solving. 
In addition, we include three multimodal datasets, including MME RealWorld (MMEWorld)~\citep{zhang2024mme}, Geometry3K (GEO3K)~\citep{lu2021inter}, and MMK12~\citep{meng2025mm}, covering tasks from perception-focused to advanced visual reasoning, ensuring a broad range of reasoning difficulties. 
The detailed dataset distribution is provided in Table~\ref{tab:data_dist}. 
We provide further details of dataset collection and preprocessing in Appendix~\ref{sec: appendix-dataset}.

\begin{table}[t]
\caption{Data distribution of training set in two stages.}
\vspace{-0.5em}
\label{tab:data_dist}
\input{tables/sft_dataset}
\vspace{-0.5em}
\end{table}

\subsection{Supervised Fine-Tuning}
To enable the model to quickly grasp the five reasoning formats and facilitate parsing in the subsequent reinforcement learning stage, we adopt SFT as our cold-start strategy.  
Specifically, we train the backbone model on the dataset constructed in Stage 1 using the standard auto-regressive objective, formally defined as:
\begin{equation}
\small
p(Y \mid X) = \prod_{i=1}^{|Y|} p_{\theta}(y_i \mid X_{\text{t},<i}),
\end{equation}
where \( \theta \) denotes the trainable parameters, and \( X \) and \( Y \) represent the input and output token sequences, respectively.
However, the model trained via SFT does not learn to select reasoning formats adaptively based on task requirements.  
The SFT model often relies excessively on direct reasoning across tasks, rather than engaging in informed decision-making (see Appendix~\ref{sec: appendix-reasoning-format-dist}).

\subsection{Length-aware Reinforcement Learning}
Reinforcement Learning has demonstrated its effectiveness in enhancing the reasoning capabilities of language models~\citep{guo2025deepseek}, often yielding better generalization than SFT.
However, traditional RL algorithms tend to constrain exploration due to reward functions focused on a single objective, such as accuracy.
While this design may be acceptable in certain contexts, it can lead to the format collapse problem when applying Group Relative Policy Optimization (GRPO)~\cite{wu2025arm}.
Specifically, format collapse refers to GRPO disproportionately favoring the reasoning format with the highest accuracy, thereby discouraging exploration of other formats, which results in excessive and unnecessary token usage.
For example, applying \formatFour{} indiscriminately for all tasks is inefficient, especially for perception-oriented tasks that do not require advanced reasoning.

To address this limitation, we propose \method{}, a novel extension of GRPO that promotes diversity in reasoning formats to mitigate format collapse, while incorporating a length-aware reward penalty. 
By explicitly integrating response length into the optimization objective, \method{} enables the model to balance reasoning accuracy with computational efficiency, thereby supporting more adaptive and cost-effective format selection.

Following ARM~\citep{wu2025arm}, \method{} incorporates a format encouragement factor to amplify the reward \(r_i\) corresponding to the response \(o_i\) for less frequently sampled reasoning formats, preventing their disappearance and ensuring adequate learning. 
Formally, we scale the original reward \(r_i\) (a binary value indicating accuracy, either 0 or 1) to \(r_i'\) as follows:

{
\small
\begin{equation}
r_i' = \alpha_i(t) \cdot r_i = \frac{G}{F(o_{i})} \cdot r_i ,
\end{equation}
}

where \(F(o_{i})\) denotes the number of times the reasoning format corresponding to \(o_i\) appears within its group \(O\).
In this way, underrepresented formats receive proportionally larger rewards, encouraging balanced exploration and mitigating the risk of format collapse.

In addition to encouraging the exploration of less frequent reasoning formats, \method{} further applies a length penalty to longer formats. 
This encourages more concise reasoning when possible and allows shorter, efficient formats to stand out when they are sufficient to solve the problem.
Formally, we further scale the reward \(r_i'\) to \(r_i''\) by:

{\small
\begin{equation}
r_i'' = \beta_i(t) \cdot r_i',
\end{equation}}

{\small
\begin{equation}
\label{eq:alfa}
\beta_i = \exp\left( - \lambda \cdot \frac{l_i - l_{\min}}{l_{\max} - l_{\min} + \epsilon} \right),
\end{equation}}

where \(l_i\) denotes the length of the response associated with \(o_i\), and \(l_{\min}\) and \(l_{\max}\) represent the minimum and maximum response lengths within the group, respectively.  
\(\lambda\) is a hyperparameter that controls the strength of the length penalty, and \(\epsilon\) is a small constant added to avoid division by zero when \(l_{\max} = l_{\min}\).

To further stabilize the reward signal throughout training, we apply a cosine decay schedule~\citep{loshchilov2017sgdr} to dynamically adjust the amplified reward. Specifically, given the intermediate reward \( r_i'' = \alpha_i \cdot \beta_i \cdot r_i \), we compute the final training reward \( \tilde{r}_i \) by modulating \( r_i'' \) with a cosine decay factor over the course of training:

{\small
\begin{equation}
\tilde{r}_i(t) = \text{Decay}(r_i'', t, T),
\end{equation}}

\noindent where \( t \) is the current training step and \( T \) is the total number of training steps. The cosine decay function is defined as:

{\small
\begin{equation}
\text{Decay}(r''_{i}, t, T) = b + 0.5 \cdot (r''_{i} - b) \cdot \left(1 + \cos\left(\pi \cdot \frac{t}{T}\right)\right),
\end{equation}
}

\noindent where \( b = 1 \) is the baseline reward to which the decayed reward converges. 
This schedule gradually reduces the influence of reward amplification as training progresses, encouraging exploration in the early stages and promoting stability and convergence in the later stages.

We adopt the standard advantage \(\hat{A}_{i,k}\) computation in GRPO, which is computed based on the group of reshaped rewards \(\tilde{r}=\{\tilde{r_1}, \tilde{r_2}, \cdots, \tilde{r_{G}}\}\):

{
\small
\begin{equation}
\label{eq:adv}
\hat{A}_{i,k} = \frac{\tilde{r_i} - \mathrm{mean}(\{\tilde{r_1}, \tilde{r_2}, \cdots, \tilde{r_{G}}\})}{\mathrm{std}(\{\tilde{r_1}, \tilde{r_2}, \cdots, \tilde{r_{G}}\})}.
\end{equation}
}

The model is finally optimized by maximizing the standard GRPO objective~\citep{shao2024deepseekmath}.

\subsection{Inference}
During inference, we set the temperature to 0.7 and the top-\(p\) value to 1.0.  
For all evaluation datasets, we use accuracy as the primary metric.  
For all reasoning formats except the code reasoning format, the final answer is directly generated by the model. 
For the code reasoning format, if the generated code is executable, we run it using an external interpreter to produce the final answer (\formatFour{}); if execution fails, the model's generated output is used instead (\formatThree{}).

%% file: tables/sft_dataset.tex
\resizebox{\linewidth}{!}{%
\begin{tabular}{@{}lccc@{}}
\toprule
Dataset & Modality & Format & Size \\
\midrule
\multicolumn{4}{c}{\textit{Supervised Finetuning}} \\
\midrule
AQuA-Rat & Text & Multiple-Choice & 3.0K \\
         & Text & Free-form & 7.8K \\
VisualWebInstruct & Text\&Vision & Multiple-Choice & 1.9K \\
                  & Text\&Vision & Free-form & 2.2K \\

\rowcolor{gray!20}
\multicolumn{3}{l}{\textbf{Total}} & 15.1K \\
\midrule
\multicolumn{4}{c}{\textit{Reinforcement Learning}} \\
\midrule
CommonsenseQA & Text & Multiple-Choice & 4.9K \\
GSM8K         & Text & Free-form & 7.4K \\
AIME         & Text & Free-form & 0.9K \\
Geometry3K   & Text\&Vision & Multiple-Choice & 2.0K \\
MME RealWorld    & Text\&Vision & Free-form & 4.0K \\
MMK12    & Text\&Vision & Free-form & 4.0K \\
\rowcolor{gray!20}
\multicolumn{3}{l}{\textbf{Total}} & 23.2K \\
\bottomrule
\end{tabular}
}

%% file: 040experiments.tex
\subsection{Experimental Setup}
\paragraph{Model}
We adopt Qwen-2.5-VL-7B~\citep{bai2025qwen2} as our backbone model due to its strong performance and demonstrated potential in reinforcement learning. 
In addition to the base model, we also evaluate Mimo-7B~\citep{coreteam2025mimovltechnicalreport}, including both its SFT and RL versions, to further examine the effectiveness of \method{} on different reasoning models. 
A detailed analysis of the impact of model choice is provided in Section~\ref{sec:backbone}.

\paragraph{Baselines}
We compare \ours{} with models trained using SFT, GRPO~\citep{shao2024deepseekmath}, and Ada-GRPO (ARM)~\citep{wu2025arm} to evaluate the effectiveness of our proposed \method{}. 
For fairness, all baselines are trained on the same data as \ours{}, and the maximum response length is capped at 4,096 tokens to control inference cost.

\paragraph{Evaluation}
For evaluation, we assess the generalization capability of \ours{} using two types of test sets: in-domain (ID) and out-of-domain (OOD).
The ID sets correspond to the datasets used in RL training, most of which are standard benchmarks. 
Specifically, we merge AIME24 and AIME25 into a single AIME test set, and for MME World and MMK12, we randomly sample 2K instances from each dataset (see Appendix~\ref{sec: appendix-dataset} for details).
To further examine generalization, we select six OOD datasets distinct from the ID sets: OBQA~\citep{mihaylov2018can}, MATH500~\citep{lightman2023let}, GPQA-Diamond~\citep{rein2024gpqa} (ranging from easy to hard), BLINK~\citep{fu2024blink}, ChartQA~\citep{masry2022chartqa}, and MMMU~\citep{yue2024mmmu} (covering perception and reasoning).

%% file: 050results.tex
\input{tables/main_results}
\begin{figure*}[t]
\centering
    \includegraphics[width=\linewidth]{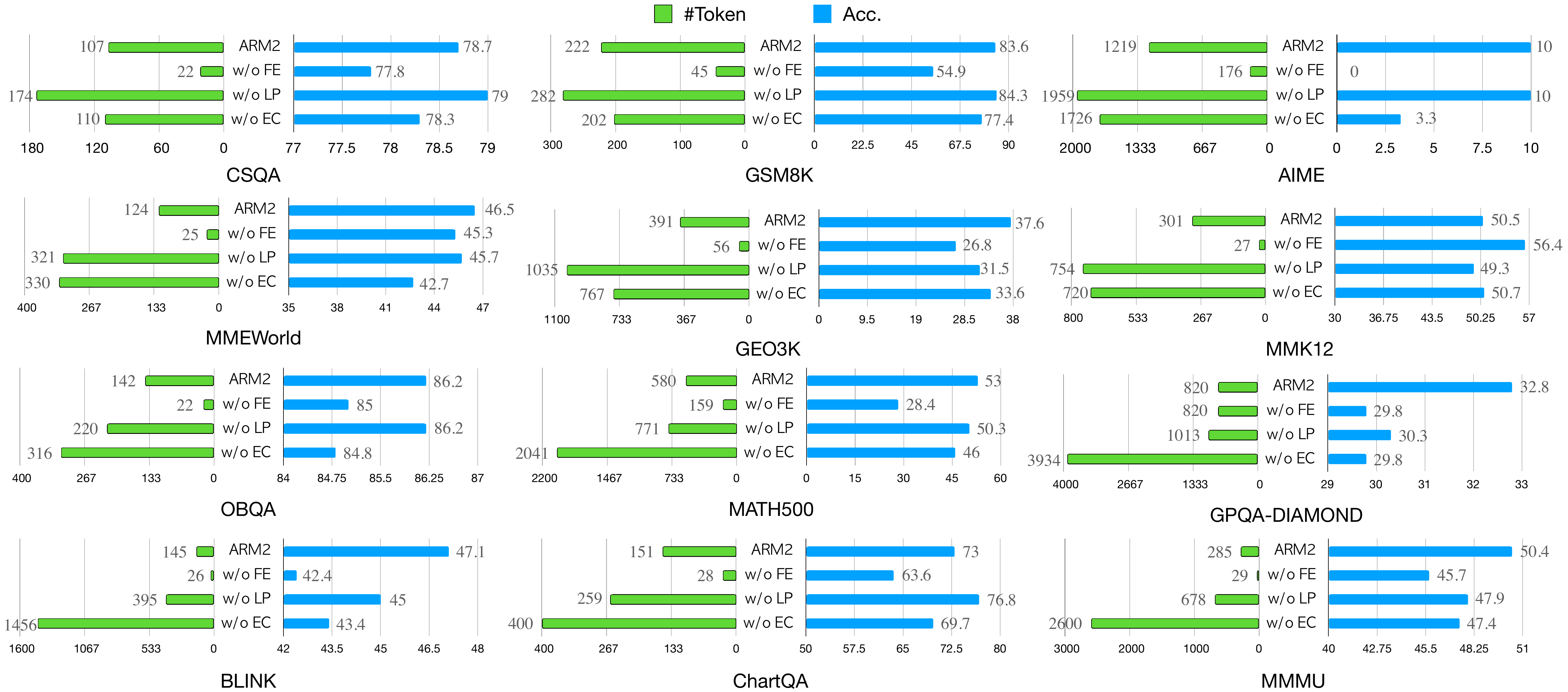}
    \caption{Ablation study of \ours{} across 12 datasets. ``w/o FE'' denotes the removal of format encouragement rewards while retaining length penalty rewards. ``w/o LP'' denotes the removal of length penalty rewards while retaining format encouragement rewards. ``w/o EC'' denotes disabling the code interpreter during both training and inference, forcing the model to reason solely in the code-text format.}
    \label{fig: ablation study}
\end{figure*}

\subsection{Main Results}
We evaluate \ours{} on both in-domain and out-of-domain datasets, demonstrating its effectiveness in balancing token efficiency and task performance, as shown in Table~\ref{tab: main_results}.
Specifically, we make the following summary based on the observation:

\paragraph{SFT and GRPO achieve task-specific strengths but lack adaptability.}
We observe that Qwen-2.5-VL-7B$_{SFT}$ and Qwen-2.5-VL-7B$_{GRPO}$ each excel on certain tasks. 
For instance, SFT performs strongly on MMEWorld, while GRPO shows advantages on GSM8K. 
However, neither model maintains consistent benefits across all benchmarks, indicating that they do not capture adaptive reasoning strategies. 
SFT consumes the fewest tokens, which allows it to perform well on tasks requiring minimal reasoning, such as MMEWorld, but it suffers substantially on more complex tasks, such as GSM8K. 
In contrast, GRPO achieves strong results on GSM8K but expends excessive tokens on easier tasks such as CSQA, and even performs worse than the more token-efficient SFT on MMEWorld.
We further study this in Section~\ref{sucsec: dist}.

\paragraph{\ours{} learns to reason adaptively and achieves better token efficiency than the previous adaptive reasoning strategy.}
Compared with SFT and GRPO, \ours{} consistently reduces token usage by more than 70\% on average relative to GRPO, while maintaining strong performance across both in-domain and out-of-domain tasks. 
Moreover, by leveraging the length-aware \method{}, \ours{} demonstrates superior token efficiency, saving over 30\% compared with the previous strategy ARM.

\subsection{Ablation Study}
\label{sec: ablation}
To evaluate the effectiveness of each module, we conduct ablation experiments by removing it individually and measuring the performance of the resulting models. 
Specifically, we remove the format encouragement reward (``w/o FE''), the length penalty reward (``w/o LP''), and executable code (``w/o EC'') during training and inference, respectively. The results are reported in Figure~\ref{fig: ablation study}. 
Our findings are summarized as follows:
\begin{inparaenum}[\it 1)]
\item \textbf{Without format encouragement, models tend to generate extremely short responses, resulting in performance degradation.}
In this setting, the length penalty reward disproportionately penalizes longer outputs, which significantly harms performance and leads to large drops across almost all tasks.
\item \textbf{Without the length penalty, responses become substantially longer, yet performance does not improve and may even decline.}
As shown in Figure~\ref{fig: ablation study}, removing the length penalty increases average token usage significantly, but yields no corresponding gains in performance.
\item \textbf{Without code execution, both performance decreases and token usage increase.}
We find that code execution has a positive effect, boosting performance and substantially reducing token costs, especially in the OOD datasets.
We attribute this to the fact that, with reliable code execution, tasks requiring heavy reasoning can be offloaded to code, reducing the need for \formatFive{} and thus improving token efficiency. 
We provide a detailed study in Section~\ref{sec: exec code}.
\end{inparaenum}
In summary, each module offers distinct advantages, collectively contributing to enhanced performance and efficiency.

%% file: tables/main_results.tex
\definecolor{caribbeangreen}{rgb}{0.0, 0.8, 0.6}
\definecolor{reddishcomplement}{rgb}{1.0, 0.2, 0.4}
\definecolor{myblue}{rgb}{0.1, 0.1, 1.0}
\newcolumntype{B}{>{\columncolor{myblue!0.5}}c}
\newcolumntype{G}{>{\columncolor{caribbeangreen!0.5}}c}

\begin{table*}[t]
\caption{Performance of various models on in-domain and out-of-domain evaluation datasets. For consistency with the training setup and to respect computational limits, the maximum pixel size is set to 512×28×28. For parsing, we adopt Qwen-2.5-14B-Instruct as the LLM judge. ``\textdagger'' indicates the test set is a processed subset. }
\setlength{\tabcolsep}{3pt}
\centering
\resizebox{\linewidth}{!}{%
\begin{tabular}{l *{7}{B}|*{7}{G}}

\toprule
  & \multicolumn{7}{c|}{\textbf{Accuracy (\(\uparrow\))}} & \multicolumn{7}{c}{\textbf{\#Tokens (\(\downarrow\))}} \\
\midrule
\rowcolor{gray!20}
\multicolumn{15}{c}{In-Domain} \\
\midrule
\multirow{3}{*}{Models} & \multicolumn{3}{c}{Text} & \multicolumn{3}{c}{Text \& Vision} & \multicolumn{1}{c|}{\multirow{2}{*}{\cellcolor{white}Avg.}}
  & \multicolumn{3}{c}{Text} & \multicolumn{3}{c}{Text \& Vision} & \multirow{2}{*}{\cellcolor{white}\phantom{A}Avg.\phantom{A}} \\
\cmidrule(lr){2-4} \cmidrule(lr){5-7} \cmidrule(lr){9-11} \cmidrule(lr){12-14}
& {\cellcolor{white}CSQA} 
& {\cellcolor{white}GSM8K} 
& {\cellcolor{white}AIME} 
& {\cellcolor{white}MMEWorld\textdagger} 
& {\cellcolor{white}GEO3K} 
& {\cellcolor{white}MMK12} 
& {}
& {\cellcolor{white}CSQA} 
& {\cellcolor{white}GSM8K} 
& {\cellcolor{white}AIME} 
& {\cellcolor{white}MMEWorld} 
& {\cellcolor{white}GEO3K} 
& {\cellcolor{white}MMK12} 
& {} \\
\midrule
Qwen-2.5-VL-7B  
  & 70.2 & 86.3 & 3.3 & 27.3 & 35.6 & 42.7 
  & 44.2 &
  180 & 291 & 1050 & 138 & 376 & 496 
  & 422 \\

\midrule
Qwen-2.5-VL-7B$_{SFT}$  
  & 75.6 & 33.3 & 0 & 48.2 & 22.8 & 53.6 
  & 38.9
  & 51 & 53 & 92 & 114 & 129 & 87 
  & 88 \\
Qwen-2.5-VL-7B$_{GRPO}$ 
  & 79.0 & 87.5 & 10.0 & 43.4 & 40.3 & 50.7 
  & 51.8 & 
  633 & 636 & 3729 & 557 & 1333 & 1462 
  & 1392 \\
\textsc{Arm}-7B
  & 79.0 & 84.3 & 10.0 & 45.7 & 31.5 & 49.3 
  & 50.0 &
  174 & 282 & 1959 & 321 & 1035 & 754 
  & 754 \\
\ours{}-7B
  & 78.7 & 83.6 & 10.0  & 46.5 & 37.6 
  & 50.5 & 51.2 &
  107 & 222 & 1219 & 124 & 391 & 301 
  & 394 \\
$\Delta$ (\ours{} $-$ Qwen-2.5-VL-7B$_{GRPO}$)
  & {\textcolor{reddishcomplement}{-\num{0.3}}} & {\textcolor{reddishcomplement}{-\num{3.9}}} & \num{0}
  & {\textcolor{caribbeangreen}{+\num{3.1}}} & {\textcolor{reddishcomplement}{-\num{2.7}}} & {\textcolor{reddishcomplement}{-\num{0.2}}} 
  & {\textcolor{reddishcomplement}{-\num{0.6}}}
  & {\textcolor{caribbeangreen}{-83.1\%}} & {\textcolor{caribbeangreen}{-65.1\%}} & {\textcolor{caribbeangreen}{-67.3\%}} 
  & {\textcolor{caribbeangreen}{-77.7\%}} & {\textcolor{caribbeangreen}{-70.7\%}} & {\textcolor{caribbeangreen}{-79.4\%}} 
  & {\textcolor{caribbeangreen}{-71.7\%}} \\

\midrule
\rowcolor{gray!20}
\multicolumn{15}{c}{Out-of-Domain} \\
\midrule

\multirow{3}{*}{Models} & \multicolumn{3}{c}{Text} & \multicolumn{3}{c}{Text \& Image} & \multicolumn{1}{c|}{\multirow{2}{*}{\cellcolor{white}Avg.}}
  & \multicolumn{3}{c}{Text} & \multicolumn{3}{c}{Text \& Image} & \multirow{2}{*}{\cellcolor{white}\phantom{A}Avg.\phantom{A}} \\
\cmidrule(lr){2-4} \cmidrule(lr){5-7} \cmidrule(lr){9-11} \cmidrule(lr){12-14}
& {\cellcolor{white}OBQA} 
& {\cellcolor{white}MATH500} 
& {\cellcolor{white}GPQA-Diamond} 
& {\cellcolor{white}BLINK} 
& {\cellcolor{white}ChartQA} 
& {\cellcolor{white}MMMU} 
& {}
& {\cellcolor{white}OBQA} 
& {\cellcolor{white}MATH500} 
& {\cellcolor{white}GPQA-Diamond} 
& {\cellcolor{white}BLINK} 
& {\cellcolor{white}ChartQA} 
& {\cellcolor{white}MMMU} 
& {} \\
\midrule
Qwen-2.5-VL-7B  
  & 66.6 & 60.8 & 25.8 & 41.6 & 67.9 & 32.4
  & 49.2 &
  149 & 561 & 616 & 134 & 106 & 341 
  & 318 \\
\midrule
Qwen-2.5-VL-7B$_{SFT}$  
  & 85.7 & 24.4 & 32.2 & 42.2 & 67.0 & 45.4 
  & 49.5
  & 51 & 78 & 84 & 135 & 81 & 105 
  & 89 \\
Qwen-2.5-VL-7B$_{GRPO}$ 
  & 85.0 & 56.4 & 32.7 & 44.8  & 73.2 & 49.2 
  & 57.1 & 
  525 & 2367 & 3569 & 728 & 553 & 1460 
  &  1534 \\
\textsc{Arm}-7B  
  & 86.2 & 50.3 & 30.3 & 45.0 & 76.8 & 47.9 
  & 56.1 & 220 & 771 & 1013 & 395 & 259 & 678 & 556 \\
\ours{}-7B  
  & 86.2 & 53.0 & 32.8  & 47.1 & 73.0 
  & 50.4 & 57.1 &
  142 & 580 & 820 & 145 & 151 & 285 
  & 354 \\
$\Delta$ (\ours{} $-$ Qwen-2.5-VL-7B$_{GRPO}$)
  & {\textcolor{caribbeangreen}{+\num{+1.2}}} & {\textcolor{reddishcomplement}{-\num{3.4}}} & {\textcolor{caribbeangreen}{+\num{0.1}}} 
  & {\textcolor{caribbeangreen}{+\num{2.3}}} & {\textcolor{reddishcomplement}{-\num{0.2}}} & {\textcolor{caribbeangreen}{+\num{1.2}}} 
  & 0
  & {\textcolor{caribbeangreen}{-73.0\%}} & {\textcolor{caribbeangreen}{-75.5\%}} & {\textcolor{caribbeangreen}{-77.0\%}} 
  & {\textcolor{caribbeangreen}{-80.1\%}} & {\textcolor{caribbeangreen}{-72.7\%}} & {\textcolor{caribbeangreen}{-80.5\%}} 
  & {\textcolor{caribbeangreen}{-77.6\%}} \\

\bottomrule
\end{tabular}%
}
\label{tab: main_results}
\end{table*}

%% file: 060analysis.tex
\begin{figure}[t]
\centering
    \includegraphics[width=\linewidth]{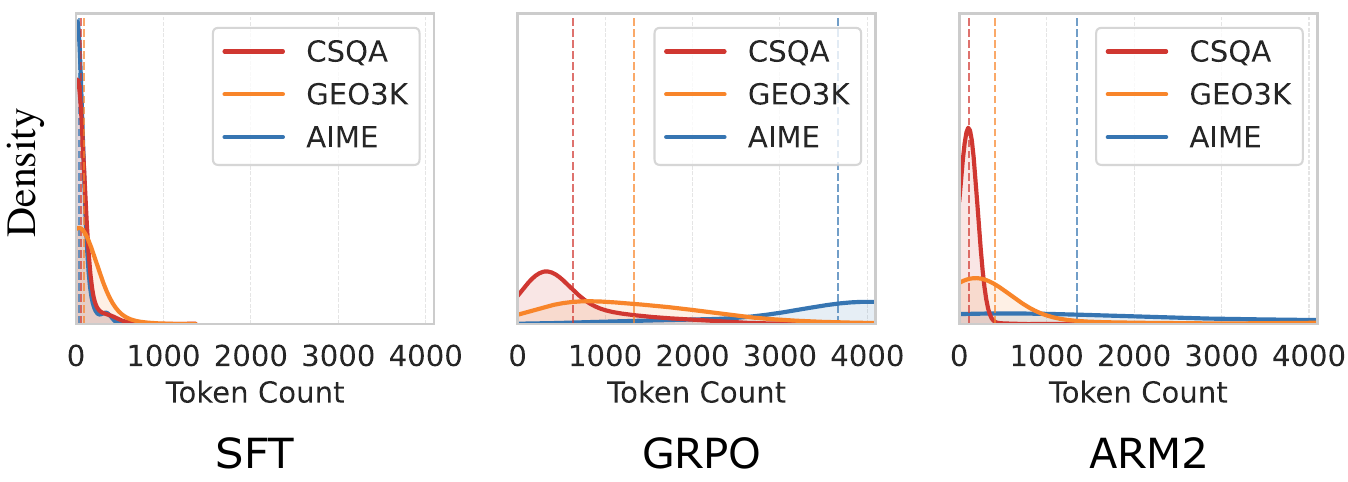}
    \vspace{-1.6em}
    \caption{Length distribution of different models across three representative datasets.  The dashed vertical line indicates the average token cost for each dataset.}
    \label{fig: length_dist}
    \vspace{-1em}
\end{figure}

\subsection{Distribution}
\label{sucsec: dist}
To illustrate why \ours{} reduces token usage and how it adaptively adjusts reasoning formats across tasks, we present the length distributions of Qwen-2.5-VL-7B$_{SFT}$, Qwen-2.5-VL-7B$_{GRPO}$, and \ours{} on three representative datasets in Figure~\ref{fig: length_dist}. 
These datasets, CSQA, GEO3K, and AIME, cover easy, medium, and hard difficulty levels, respectively. 
We observe that the distribution of Qwen-2.5-VL-7B$_{SFT}$ is concentrated at lower token costs, while Qwen-2.5-VL-7B$_{GRPO}$ exhibits a more dispersed distribution with higher token costs. 
In contrast, as task difficulty increases, the peak of \ours{} gradually shifts toward higher token counts, demonstrating its ability to allocate longer reasoning adaptively when needed.
The reasoning format distribution is provided in Appendix~\ref{sec: appendix-reasoning-format-dist}.

\subsection{Test-Time Scaling}
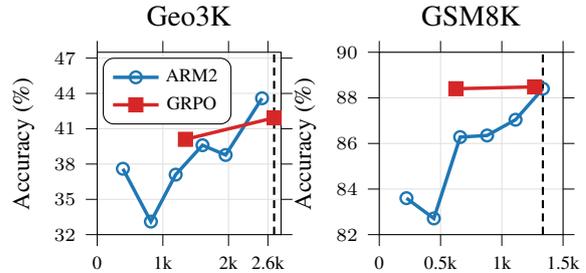
\begin{figure}[t]
\vspace{-2em}
\centering
\begin{subfigure}{0.44\columnwidth}
  \centering
  \input{tables/tts_geo3k.tex}
\end{subfigure}
\hspace{0.017\columnwidth}
\begin{subfigure}{0.44\columnwidth}
  \centering
  \input{tables/tts_gsm8k.tex}
\end{subfigure}
\vspace{-1.5em}
\caption{\ours{} vs GRPO with varying token budgets.}
\vspace{-0.5em}
\label{fig:tts}
\end{figure}

Test-time scaling enables models to allocate additional tokens at inference to deliver a more reliable answer. 
We investigate whether \ours{} can benefit from this strategy, particularly under the same token budget as the original reasoning models. 
Our analysis focuses on two reasoning-intensive tasks, GSM8K and GEO3K, where \ours{} initially lags behind GRPO.
We implement test-time scaling via majority voting~\citep{wang2023self}: sampling multiple responses and adopting the most frequent one as the final prediction. 
To ensure fairness, we constrain the overall token budget and compare performance accordingly. 
The results are presented in Figure~\ref{fig:tts}.
Our findings show that, although \ours{} exhibits a slight performance drop at lower token budgets (e.g., on GSM8K and GEO3K), test-time scaling effectively mitigates this gap. 
Moreover, as the token budget increases, \ours{} even surpasses GRPO under the same budget, demonstrating that \method{} not only improves token efficiency but also preserves reasoning capability when sufficient budget is available.

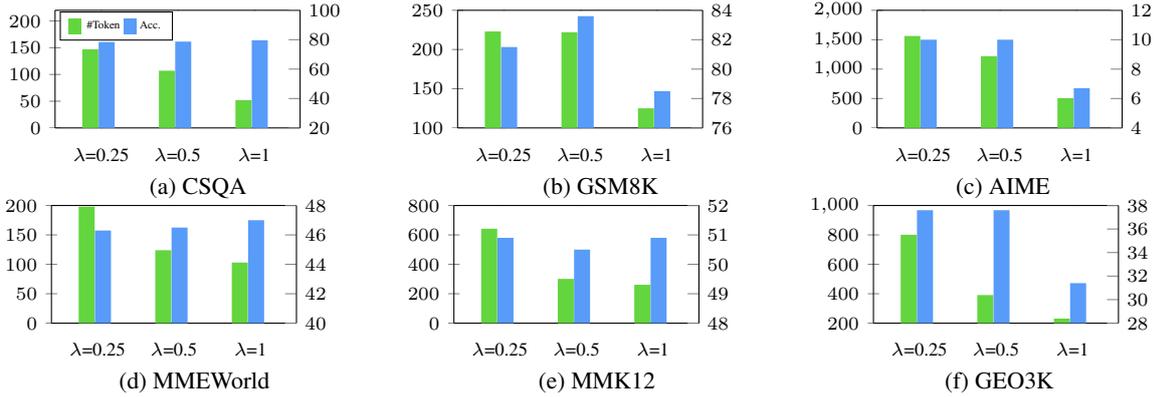
\begin{figure*}[t]
    \centering
    \begin{subfigure}{.325\textwidth}
        \input{tables/csqa_lp}
        \vspace{-0.4em}
        \caption{CSQA}
    \end{subfigure}
    \begin{subfigure}{.325\textwidth}
        \input{tables/gsm8k_lp}
        \vspace{-0.4em}
        \caption{GSM8K}
    \end{subfigure}
    \begin{subfigure}{.325\textwidth}
        \input{tables/aime_lp}
        \vspace{-0.4em}
        \caption{AIME}
    \end{subfigure}
    
    \vspace{-0.5em} 

    \begin{subfigure}{.325\textwidth}
        \input{tables/mme_lp}
        \vspace{-0.4em}
        \caption{MMEWorld}
    \end{subfigure}
    \begin{subfigure}{.325\textwidth}
        \input{tables/mmk12_lp}
        \vspace{-0.4em}
    \caption{MMK12}
    \end{subfigure}
    \begin{subfigure}{.325\textwidth}
        \input{tables/geo3k_lp}
        \vspace{-0.4em}
        \caption{GEO3K}
    \end{subfigure}
    \vspace{-0.5em}
    \caption{Performance of \ours{} under different length penalty strengths.}
    \vspace{-1em}
    \label{fig: length_penalty}
\end{figure*}

\begin{figure}[t]
\centering
    \includegraphics[width=\linewidth]{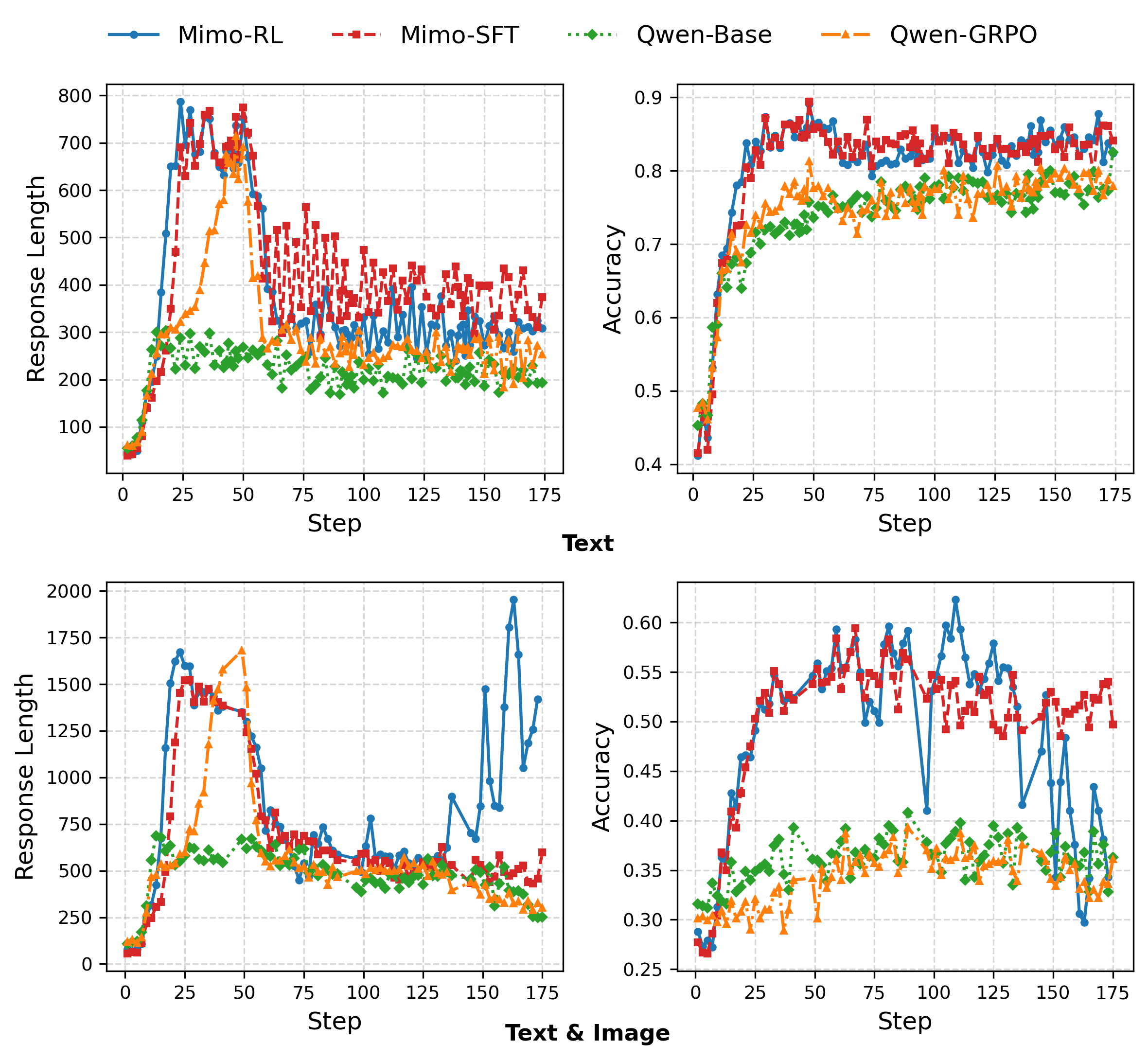}
    \vspace{-1.8em}
    \caption{Training trends of \ours{} on text and multimodal datasets with different backbone models.}
    \label{fig: training trend}
    \vspace{-1.5em}
\end{figure}

\input{tables/exec_code}

\subsection{The Effect of Executable Code}
\label{sec: exec code}
In Section~\ref{sec: ablation}, we show that incorporating executable code can substantially reduce token costs while simultaneously improving performance.
We hypothesize that executable code enhances the accuracy of code reasoning and thereby substitutes for lengthy chain-of-thought reasoning.
To validate this, we examine the proportion and accuracy of code reasoning before and after introducing the code interpreter across three OOD datasets, as reported in Table~\ref{tab:exec code}.
We observe that both the proportion of code and its accuracy increase with the use of executable code, confirming its effectiveness.

\subsection{Length Penalty Strength}
We further examine the effect of different values of $\lambda$, which controls the strength of the length penalty. Specifically, we vary $\lambda$ from \num{0.25} to \num{1} and report the results in Figure~\ref{fig: length_penalty}. 
The results reveal that the optimal response length is highly task dependent. 
For relatively easy or perception-oriented tasks such as CSQA and MMEWorld, shorter outputs tend to perform better, since they reduce redundancy, avoid over-reasoning, and align with the fact that these tasks typically require only shallow reasoning or factual recognition. 
However, on reasoning-intensive tasks such as GSM8K and MMK12, stronger length penalties harm performance, since the models truncate reasoning chains prematurely and fail to capture the multi-step logic needed for correct answers.
To balance performance and efficiency, we set $\lambda$ as \num{0.5} for all model training in this paper unless otherwise specified.

\subsection{Backbone models}
\label{sec:backbone}
For \ours{}, we further investigate the effect of different backbone initializations. 
Specifically, we compare Mimo-RL and Mimo-SFT~\citep{coreteam2025mimovltechnicalreport}, which are initialized with RL and SFT, respectively. 
In addition, we evaluate our own GRPO-initialized Qwen backbone, which is trained with GRPO for 50 steps on the same RL datasets.

The training dynamics of accuracy and response length for these models are illustrated in Figure~\ref{fig: training trend}. Our findings suggest several key observations:
\begin{inparaenum}[\it 1)]
\item 
\textbf{Stronger backbones yield substantial gains.}
On the training set, Mimo-based models achieve nearly a 10\% improvement in accuracy compared to the weaker model Qwen, highlighting the importance of initialization quality.
\item 
\textbf{Instability from RL initialization}
Despite its higher starting performance, Mimo-RL exhibits noticeable oscillations when trained on multimodal datasets. 
This instability ultimately reduces its effectiveness and suggests that continual fine-tuning from an RL backbone may amplify variance in cross-modal learning.
\item
\textbf{Limited gains from GRPO initialization. }
The comparison between models with and without GRPO initialization shows only marginal differences, indicating that lightweight GRPO pretraining alone does not provide a decisive advantage for \ours{}. 
Greater benefits arise when GRPO is combined with large-scale post-training, as demonstrated in Mimo-RL.
\end{inparaenum}

%% file: tables/tts_geo3k.tex
\begin{tikzpicture}
\definecolor{ARblue}{HTML}{1F77B4}
\definecolor{GRred}{HTML}{D62728}
\def\cutB{2694.9} 

\begin{axis}[
  width=4cm, height=4cm,
  title={Geo3K},
  ylabel={Accuracy (\%)},
  xlabel style={font=\footnotesize},
  ylabel style={font=\footnotesize},
  ymin=32, ymax=47.5, ytick={32,35,38,41,44,47},
  xmin=0, xmax=2800,              
  xtick={0,1000,2000,2600},
  xticklabels={0,1k,2k,2.6k},
  grid=both, grid style={draw=gray!25, line width=.3pt},
  minor grid style={draw=gray!18},
  tick style={black}, tick align=outside,
  tick label style={font=\scriptsize},
  every axis plot/.style={line width=1.3pt, mark size=2.2pt},
  legend style={
    draw=black, fill=white, rounded corners,
    font=\scriptsize, at={(0.03,0.97)}, anchor=north west
  },
]

\addplot+[ARblue, mark=o, mark options={fill=white, line width=.9pt}] coordinates {
  (391,37.6) (817.6,33.11) (1195.2,37.10)
  (1604.1,39.60) (1956.1,38.77) (2510.8,43.59)
};
\addlegendentry{ARM2}

\addplot+[GRred, mark=square*, mark options={solid}] coordinates {
  (1344.4,40.10) (2694.9,41.93)
};
\addlegendentry{GRPO}

\addplot [densely dashed, black, line width=.8pt]
  coordinates {(\cutB,32) (\cutB,47.5)};
\node[font=\footnotesize, anchor=south] at (axis cs:\cutB,47.5) {cutoff};
\end{axis}
\end{tikzpicture}

%% file: tables/tts_gsm8k.tex
\begin{tikzpicture}
\definecolor{ARblue}{HTML}{1F77B4}
\definecolor{GRred}{HTML}{D62728}
\def\cutA{1334.9} 

\begin{axis}[
  width=4cm, height=4cm,
  title={GSM8K},
ylabel={Accuracy (\%)},
  xlabel style={font=\footnotesize},
  ylabel style={font=\footnotesize},
  ymin=82, ymax=90, ytick={82,84,86,88,90},
  xmin=0, xmax=1500,               
  xtick={0,500,1000,1500},
  xticklabels={0,0.5k,1k,1.5k},    
  grid=both, grid style={draw=gray!25, line width=.3pt},
  minor grid style={draw=gray!18},
  tick style={black}, tick align=outside,
  tick label style={font=\scriptsize},
  every axis plot/.style={line width=1.3pt, mark size=2.2pt},
  legend style={
    draw=black, fill=white, rounded corners,
    font=\scriptsize, at={(0.03,0.97)}, anchor=north west
  },
]

\addplot+[ARblue, mark=o, mark options={fill=white, line width=.9pt}] coordinates {
  (222,83.6) (445.2,82.71) (660.3,86.28)
  (879.9,86.35) (1111.9,87.04) (1334.9,88.40)
};

\addplot+[GRred, mark=square*, mark options={solid}] coordinates {
  (626.0,88.40) (1267.8,88.48)
};

\addplot [densely dashed, black, line width=.8pt]
  coordinates {(\cutA,82) (\cutA,90)};
\node[font=\footnotesize, anchor=south] at (axis cs:\cutA,90) {cutoff};
\end{axis}
\end{tikzpicture}

%% file: tables/csqa_lp.tex
\definecolor{CustomColor1}{RGB}{99, 213, 63}
\definecolor{CustomColor2}{RGB}{88, 155, 249}

\pgfplotsset{width=0.62\linewidth,height=0.3\linewidth,compat=1.5,scale only axis}

\footnotesize

\begin{tikzpicture}

 \begin{axis}[
        ybar,
        bar width=6pt,
        bar shift=-4pt,
        symbolic x coords={$\lambda$=0.25, $\lambda$=0.5, $\lambda$=1},
        ymin=0, ymax=220,
        enlarge x limits=0.3,
        scaled ticks=false,
        legend pos=north west,
        xtick style={/pgfplots/major tick length=0pt},
        legend style={nodes={scale=0.7, transform shape},font=\scriptsize},
        xticklabel style={font=\scriptsize},
        yticklabel style={font=\scriptsize},
        axis y line=left,  
        axis line style={-},
        legend style={           
        draw=black,  
            font=\tiny,
            at={(0.25,0.99)},
            anchor=north,  
            legend columns=2, 
            /tikz/column 1/.style={column sep=1pt} 
        },
        legend image code/.code={%
           \draw[fill, draw=none] (0cm,-0.1cm) rectangle (0.16cm,0.1cm);
        }
    ]
        \addplot+[fill=CustomColor1, draw=none] coordinates  {
            ($\lambda$=0.25, 147)
            ($\lambda$=0.5, 107)
            ($\lambda$=1, 52)

        };
        \addlegendentry{\#Token}

        \addplot+[fill=CustomColor2, draw=none] coordinates  {
            ($\lambda$=0.25, 0)
            ($\lambda$=0.5, 0)
            ($\lambda$=1, 0)

        };
        
        \addlegendentry{Acc.}
        
    \end{axis}

  \begin{axis}[
    ybar,
    bar width=6pt,
    bar shift=+2.1pt,   
    symbolic x coords={$\lambda$=0.25, $\lambda$=0.5, $\lambda$=1},
    ymin=20, ymax=100,
    enlarge x limits=0.3,
    axis y line*=right,
    axis x line=none,
    ytick pos=right,
    yticklabel style={font=\scriptsize},
    legend style={at={(0.8,0.9)},anchor=north,draw=none,font=\tiny}
  ]
    \addplot+[fill=CustomColor2, draw=none] coordinates {
      ($\lambda$=0.25, 78.4)
      ($\lambda$=0.5, 78.7)
      ($\lambda$=1, 79.6)
    };
    
  \end{axis}
\end{tikzpicture}

%% file: tables/gsm8k_lp.tex
\definecolor{CustomColor1}{RGB}{99, 213, 63}
\definecolor{CustomColor2}{RGB}{88, 155, 249}

\pgfplotsset{width=0.62\linewidth,height=0.3\linewidth,compat=1.5,scale only axis}

\footnotesize

\begin{tikzpicture}

 \begin{axis}[
        ybar,
        bar width=6pt,
        bar shift=-4pt,
        symbolic x coords={$\lambda$=0.25, $\lambda$=0.5, $\lambda$=1},
        ymin=100, ymax=250,
        enlarge x limits=0.3,
        scaled ticks=false,
        legend pos=north west,
        xtick style={/pgfplots/major tick length=0pt},
        legend style={nodes={scale=0.7, transform shape},font=\scriptsize},
        xticklabel style={font=\scriptsize},
        yticklabel style={font=\scriptsize},
        axis y line=left,   
        axis line style={-},
        legend style={           
        draw=black,  
            font=\tiny,
            at={(0.25,0.99)},
            anchor=north,  
            legend columns=2, 
            /tikz/column 1/.style={column sep=1pt} 
        },
        legend image code/.code={%
           \draw[fill, draw=none] (0cm,-0.1cm) rectangle (0.16cm,0.1cm);
        }
    ]
        \addplot+[fill=CustomColor1, draw=none] coordinates  {
            ($\lambda$=0.25, 223)
            ($\lambda$=0.5, 222)
            ($\lambda$=1, 125)

        };

    \end{axis}

  \begin{axis}[
    ybar,
    bar width=6pt,
    bar shift=+2.1pt, 
    symbolic x coords={$\lambda$=0.25, $\lambda$=0.5, $\lambda$=1},
    ymin=76, ymax=84,
    enlarge x limits=0.3,
    axis y line*=right,
    axis x line=none,
    ytick pos=right,
    yticklabel style={font=\scriptsize},
    legend style={at={(0.8,0.9)},anchor=north,draw=black,font=\tiny}
  ]
    \addplot+[fill=CustomColor2, draw=none] coordinates {
      ($\lambda$=0.25, 81.5)
      ($\lambda$=0.5, 83.6)
      ($\lambda$=1, 78.5)
    };
    
  \end{axis}
\end{tikzpicture}

%% file: tables/aime_lp.tex
\definecolor{CustomColor1}{RGB}{99, 213, 63}
\definecolor{CustomColor2}{RGB}{88, 155, 249}

\pgfplotsset{width=0.62\linewidth,height=0.3\linewidth,compat=1.5,scale only axis}

\footnotesize

\begin{tikzpicture}

 \begin{axis}[
        ybar,
        bar width=6pt,
        bar shift=-4pt,
        symbolic x coords={$\lambda$=0.25, $\lambda$=0.5, $\lambda$=1},
        ymin=0, ymax=2000,
        enlarge x limits=0.3,
        scaled ticks=false,
        legend pos=north west,
        xtick style={/pgfplots/major tick length=0pt},
        legend style={nodes={scale=0.7, transform shape},font=\scriptsize},
        xticklabel style={font=\scriptsize},
        yticklabel style={font=\scriptsize},
        axis y line=left,   
        axis line style={-},
        legend style={           
        draw=black,  
            font=\tiny,
            at={(0.25,0.99)},
            anchor=north,  
            legend columns=2, 
            /tikz/column 1/.style={column sep=1pt} 
        },
        legend image code/.code={%
           \draw[fill, draw=none] (0cm,-0.1cm) rectangle (0.16cm,0.1cm);
        }
    ]
        \addplot+[fill=CustomColor1, draw=none] coordinates  {
            ($\lambda$=0.25, 1563)
            ($\lambda$=0.5, 1219)
            ($\lambda$=1, 506)

        };

    \end{axis}

  \begin{axis}[
    ybar,
    bar width=6pt,
    bar shift=+2.1pt,   
    symbolic x coords={$\lambda$=0.25, $\lambda$=0.5, $\lambda$=1},
    ymin=4, ymax=12,
    enlarge x limits=0.3,
    axis y line*=right,
    axis x line=none,
    ytick pos=right,
    yticklabel style={font=\scriptsize},
    legend style={at={(0.8,0.9)},anchor=north,draw=black,font=\tiny}
  ]
    \addplot+[fill=CustomColor2, draw=none] coordinates {
      ($\lambda$=0.25, 10.0)
      ($\lambda$=0.5, 10.0)
      ($\lambda$=1, 6.7)
    };
    
  \end{axis}
\end{tikzpicture}

%% file: tables/mme_lp.tex
\definecolor{CustomColor1}{RGB}{99, 213, 63}
\definecolor{CustomColor2}{RGB}{88, 155, 249}

\pgfplotsset{width=0.62\linewidth,height=0.3\linewidth,compat=1.5,scale only axis}

\footnotesize

\begin{tikzpicture}

 \begin{axis}[
        ybar,
        bar width=6pt,
        bar shift=-4pt,
        symbolic x coords={$\lambda$=0.25, $\lambda$=0.5, $\lambda$=1},
        ymin=0, ymax=200,
        enlarge x limits=0.3,
        scaled ticks=false,
        legend pos=north west,
        xtick style={/pgfplots/major tick length=0pt},
        legend style={nodes={scale=0.7, transform shape},font=\scriptsize},
        xticklabel style={font=\scriptsize},
        yticklabel style={font=\scriptsize},
        axis y line=left,   
        axis line style={-},
        legend style={           
        draw=black,  
            font=\tiny,
            at={(0.25,0.99)},
            anchor=north,  
            legend columns=2, 
            /tikz/column 1/.style={column sep=1pt} 
        },
        legend image code/.code={%
           \draw[fill, draw=none] (0cm,-0.1cm) rectangle (0.16cm,0.1cm);
        }
    ]
        \addplot+[fill=CustomColor1, draw=none] coordinates  {
            ($\lambda$=0.25, 198)
            ($\lambda$=0.5, 124)
            ($\lambda$=1, 103)

        };

    \end{axis}

  \begin{axis}[
    ybar,
    bar width=6pt,
    bar shift=+2.1pt,   
    symbolic x coords={$\lambda$=0.25, $\lambda$=0.5, $\lambda$=1},
    ymin=40, ymax=48,
    enlarge x limits=0.3,
    axis y line*=right,
    axis x line=none,
    ytick pos=right,
    yticklabel style={font=\scriptsize},
    legend style={at={(0.8,0.9)},anchor=north,draw=black,font=\tiny}
  ]
    \addplot+[fill=CustomColor2, draw=none] coordinates {
      ($\lambda$=0.25, 46.3)
      ($\lambda$=0.5, 46.5)
      ($\lambda$=1, 47)
    };
    
  \end{axis}
\end{tikzpicture}

%% file: tables/mmk12_lp.tex
\definecolor{CustomColor1}{RGB}{99, 213, 63}
\definecolor{CustomColor2}{RGB}{88, 155, 249}

\pgfplotsset{width=0.62\linewidth,height=0.3\linewidth,compat=1.5,scale only axis}

\footnotesize

\begin{tikzpicture}

 \begin{axis}[
        ybar,
        bar width=6pt,
        bar shift=-4pt,
        symbolic x coords={$\lambda$=0.25, $\lambda$=0.5, $\lambda$=1},
        ymin=0, ymax=800,
        enlarge x limits=0.3,
        scaled ticks=false,
        legend pos=north west,
        xtick style={/pgfplots/major tick length=0pt},
        legend style={nodes={scale=0.7, transform shape},font=\scriptsize},
        xticklabel style={font=\scriptsize},
        yticklabel style={font=\scriptsize},
        axis y line=left,   
        axis line style={-},
        legend style={           
        draw=black,  
            font=\tiny,
            at={(0.25,0.99)},
            anchor=north,  
            legend columns=2, 
            /tikz/column 1/.style={column sep=1pt} 
        },
        legend image code/.code={%
           \draw[fill, draw=none] (0cm,-0.1cm) rectangle (0.16cm,0.1cm);
        }
    ]
        \addplot+[fill=CustomColor1, draw=none] coordinates  {
            ($\lambda$=0.25, 642)
            ($\lambda$=0.5, 301)
            ($\lambda$=1, 261)

        };

    \end{axis}

  \begin{axis}[
    ybar,
    bar width=6pt,
    bar shift=+2.1pt,   
    symbolic x coords={$\lambda$=0.25, $\lambda$=0.5, $\lambda$=1},
    ymin=48, ymax=52,
    enlarge x limits=0.3,
    axis y line*=right,
    axis x line=none,
    ytick pos=right,
    yticklabel style={font=\scriptsize},
    legend style={at={(0.8,0.9)},anchor=north,draw=black,font=\tiny}
  ]
    \addplot+[fill=CustomColor2, draw=none] coordinates {
      ($\lambda$=0.25, 50.9)
      ($\lambda$=0.5, 50.5)
      ($\lambda$=1, 50.9)
    };
    
  \end{axis}
\end{tikzpicture}

%% file: tables/geo3k_lp.tex
\definecolor{CustomColor1}{RGB}{99, 213, 63}
\definecolor{CustomColor2}{RGB}{88, 155, 249}

\pgfplotsset{width=0.62\linewidth,height=0.3\linewidth,compat=1.5,scale only axis}

\footnotesize

\begin{tikzpicture}

 \begin{axis}[
        ybar,
        bar width=6pt,
        bar shift=-4pt,
        symbolic x coords={$\lambda$=0.25, $\lambda$=0.5, $\lambda$=1},
        ymin=200, ymax=1000,
        enlarge x limits=0.3,
        scaled ticks=false,
        legend pos=north west,
        xtick style={/pgfplots/major tick length=0pt},
        legend style={nodes={scale=0.7, transform shape},font=\scriptsize},
        xticklabel style={font=\scriptsize},
        yticklabel style={font=\scriptsize},
        axis y line=left,  
        axis line style={-},
        legend style={           
        draw=black,  
            font=\tiny,
            at={(0.25,0.99)},
            anchor=north,  
            legend columns=2, 
            /tikz/column 1/.style={column sep=1pt} 
        },
        legend image code/.code={%
           \draw[fill, draw=none] (0cm,-0.1cm) rectangle (0.16cm,0.1cm);
        }
    ]
        \addplot+[fill=CustomColor1, draw=none] coordinates  {
            ($\lambda$=0.25, 801)
            ($\lambda$=0.5, 391)
            ($\lambda$=1, 231)

        };

    \end{axis}

  \begin{axis}[
    ybar,
    bar width=6pt,
    bar shift=+2.1pt,  
    symbolic x coords={$\lambda$=0.25, $\lambda$=0.5, $\lambda$=1},
    ymin=28, ymax=38,
    enlarge x limits=0.3,
    axis y line*=right,
    axis x line=none,
    ytick pos=right,
    yticklabel style={font=\scriptsize},
    legend style={at={(0.8,0.9)},anchor=north,draw=black,font=\tiny}
  ]
    \addplot+[fill=CustomColor2, draw=none] coordinates {
      ($\lambda$=0.25, 37.6)
      ($\lambda$=0.5, 37.6)
      ($\lambda$=1, 31.4)
    };
    
  \end{axis}
\end{tikzpicture}

%% file: tables/exec_code.tex
\begin{table}
\caption{The performance comparison between \ours{} with and without executable code.}
\vspace{-1em}
\resizebox{\linewidth}{!}{
\begin{tabular}{lcccc}
\toprule
        & \multicolumn{2}{c}{w/ Code-Exec} & \multicolumn{2}{c}{w/o Code-Exec} \\ \cmidrule(lr){2-3} \cmidrule(lr){4-5}
        & Prop.(\%)           & Acc.           & Prop. (\%)           & Acc.           \\ \midrule
MATH500 & \textbf{52.6}            & \textbf{47.7}           & 22.8             & 38.6           \\
ChartQA & \textbf{47.4}            & \textbf{70.3}           & 27.8             & 65.2           \\
MMMU    & \textbf{42.0}            & \textbf{50.2}           & 21.9             & 49.8           \\ \bottomrule
\end{tabular}
}
\label{tab:exec code}
\vspace{-1em}
\end{table}

%% file: 070conclusion.tex
In this paper, we propose \ours{}, a multimodal reasoning model capable of adaptive reasoning across tasks. 
\ours{} is trained with \method{}, a variant of GRPO that incorporates format encouragement and a length penalty, enabling the model to select among five distinct reasoning formats with high token efficiency. 
Furthermore, \ours{} integrates executable code into reasoning, which preserves performance while substantially reducing token usage.
Comprehensive experiments validate the effectiveness of \ours{}, and ablation studies confirm the soundness of its design. 
Overall, \ours{} offers a promising direction for adaptive reasoning by leveraging the complementary strengths of different reasoning formats.

%% file: 080appendix.tex
\subsection{Dataset Collection and Preprocessing}
\label{sec: appendix-dataset}
For CSQA, GSM8K, and GEO3K, we adopt the standard training sets for RL training and the validation/test sets for evaluation. 
For datasets without official training splits, including AIME and MMEWorld, we construct our own training and test sets. 
Specifically, for AIME, we collect problems and corresponding solutions from 1987–2023, reserving the 60 problems from 2024–2025 as the test set.
For MMEWorld, we randomly sample 4,000 instances for training and 2,000 instances for testing. 
For MMK12, we use the standard test set for evaluation and randomly sample 4,000 instances from the training set for RL training.

\subsection{Reasoning Format Distribution}
\label{sec: appendix-reasoning-format-dist}
\begin{figure}[h]
\centering
    \includegraphics[width=\linewidth]{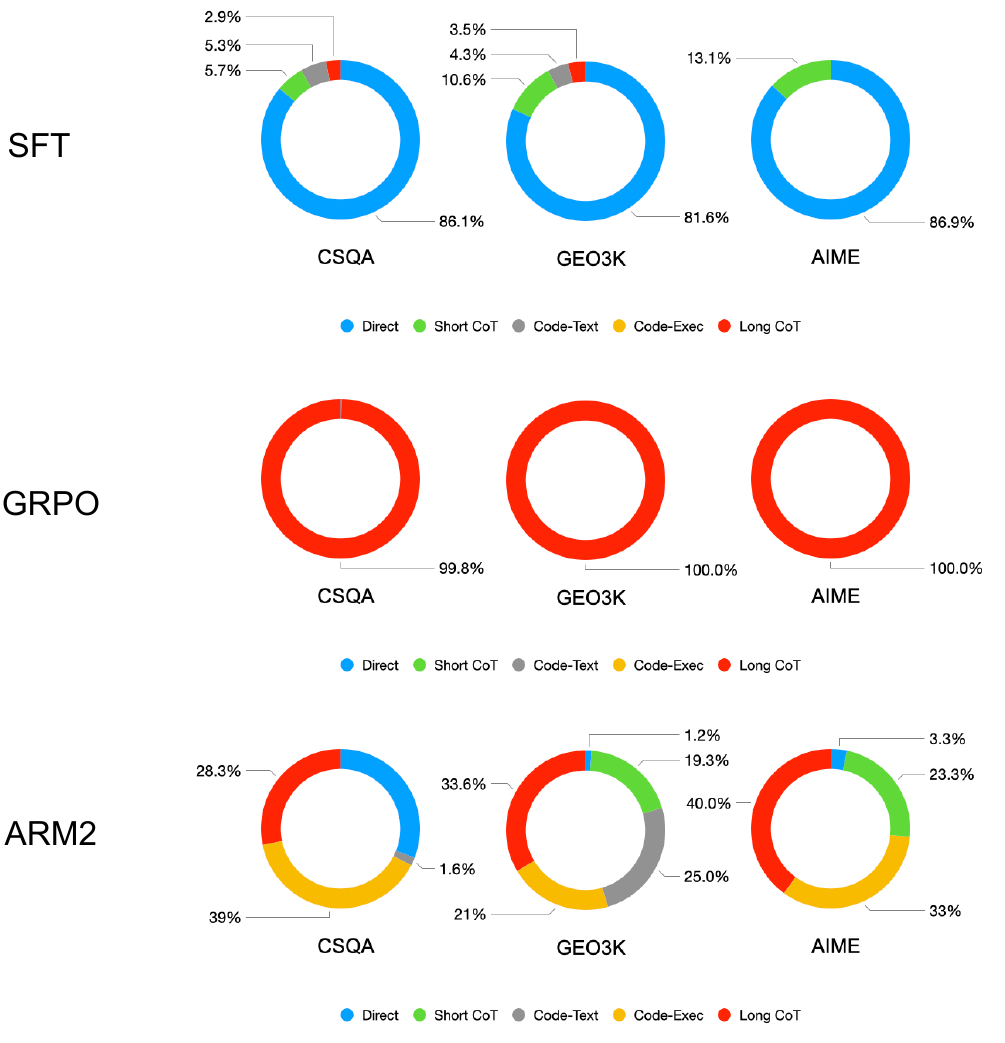}
    \caption{Reasoning format distribution.}
    \label{fig: format_dist}
    \vspace{-1em}
\end{figure}

In Figure~\ref{fig: format_dist}, we show the distribution of reasoning formats across three models and three datasets. 
The SFT model tends to rely heavily on direct reasoning for all tasks, while the GRPO model predominantly favors long chain-of-thought reasoning. 
In contrast, \ours{} demonstrates adaptive reasoning by adjusting its format selection according to task characteristics.

\subsection{Implementation Details}
\subsubsection{Data Collection}
For VisualWebInstruct, each question is paired with both an answer and a rationale (short CoT). To obtain long CoT reasoning, we prompt Doubao-V1.6-Thinking, with the prompt provided below. 
For code-based reasoning, we prompt Doubao-V1.6-Thinking to generate a function, which is then executed by an external Python interpreter. 
We retain only those cases where both the long chain-of-thought and the generated code yield the correct ground truth.

\noindent
Prompt for generating Long CoT:
\input{tables/appendix_doubao_longCoT}

\noindent
Prompt for generating Code:

\input{tables/appendix_doubao_longCoT}
\subsubsection{SFT}
We train the SFT model using LoRA~\citep{hu2022lora}, with a rank of 8. The maximum sequence length is set to 8,192 tokens, and training is performed for up to 6 epochs.
We use a learning rate of $2\times10^{-4}$ with a cosine scheduler and a warmup ratio of 0.1. 
The validation split is 0.1, sampled from the training set. 
The batch size for training is 2 per GPU, with a gradient accumulation step of 8.
The SFT model is trained through 8$\times$A100, which took $\sim$12 hours.

\subsubsection{RL}
We implement reinforcement learning using the VERL framework~\citep{sheng2024hybridflow}. The training batch size is set to 480, with a mini-batch size of 96. During the rollout stage, each instance generates 8 samples, with a temperature of 0.7 and a top-$p$ value of 1.0. The learning rate is fixed at $1\times10^{-6}$.
The RL model is trained through 8$\times$A100 for 150 steps, which took$\sim$12 hours.

\subsubsection{Data Examples}
We provide an example with four reasoning formats below.
\input{tables/appendix_example}

%% file: tables/appendix_doubao_longCoT.tex
\lstset{
    backgroundcolor=\color[RGB]{245,245,245},
    breaklines=true,
    breakindent=0pt,
    basicstyle=\ttfamily\small,
    frame=trbl,
    frameround = tttt,
}

\begin{lstlisting}
You are a helpful assistant who can help me extract the answer and rationale from the following text. Note that you should help me remove special characters and only return plain text. Except for the final answer, all the rest in the response should be regarded as rationale. Besides, you should tell me whether the question type is free-form or multiple-choice.  For a multiple-choice question, only return the answer in ['A', 'B',' C', D']. 
Remove the unit if the answer is a number. Follow the format: Answer: xxx\n\nRationale: xxx\n\nFormat: xxx

Question: <PLACEHOLDER>
\end{lstlisting}

%% file: tables/appendix_example.tex
\lstset{
    backgroundcolor=\color[RGB]{245,245,245},
    breaklines=true,
    breakindent=0pt,
    basicstyle=\ttfamily\small,
    frame=trbl,
    frameround = tttt,
}

\begin{lstlisting}
Question:
<image>
The following bullet points appear in a proposal. 
What is the most likely heading for the section these bullet points are in?
Reduce medical errors by 25% within one year of implementing the new EHR system.
Increase patient satisfaction with the EHR system by 10% within one year of implementation.
Reduce administrative costs associated with the EHR system by 15% within one year of implementation.

<ANSWER>
Project Objectives
</ANSWER>

<COT>
The bullet points outline specific, measurable goals to be achieved within a specified timeframe after implementing a new EHR system. They focus on outcomes or targets that the proposal aims to achieve, which aligns with the heading \"Project Objectives.\" The other options do not match the content of the bullet points as closely.
</COT>
<ANSWER>
Project Objectives
</ANSWER>

<CODE>
def determine_heading_for_bullet_points():
    headings = [
        "Potential Solutions",
        "Project Objectives",
        "Statement of the Problem",
        "Implementation Plan"
    ]
    bullet_points = [
        "Reduce medical errors by 25% within one year of implementing the new EHR system.",
        "Increase patient satisfaction with the EHR system by 10% within one year of implementation.",
        "Reduce administrative costs associated with the EHR system by 15% within one year of implementation."
    ]
    # Analysis:
    # The bullet points are all specific, measurable targets to be achieved after implementation.
    # This matches the definition of "Project Objectives" best.
    # They are not solutions, not a problem statement, and not an implementation plan.
    answer = "Project Objectives"
    index = headings.index(answer)
    return {
        'bullet_points': bullet_points,
        'chosen_heading': answer,
        'heading_index': index,
        'answer': answer
    }

>>> determine_heading_for_bullet_points()
</CODE>
<OBSERVATION>
output = {
    'bullet_points': [
        "Reduce medical errors by 25% within one year of implementing the new EHR system.",
        "Increase patient satisfaction with the EHR system by 10% within one year of implementation.",
        "Reduce administrative costs associated with the EHR system by 15% within one year of implementation."
    ],
    'chosen_heading': 'Project Objectives',
    'heading_index': 1,
    'answer': 'Project Objectives'
}</OBSERVATION>
<ANSWER>
Project Objectives
</ANSWER>

<LONG_COT>
Got it, let's see. The question is about finding the right heading for those bullet points. First, I need to remember what each section title means. 

The bullet points are all about specific, measurable goals: reduce errors by 25%, increase satisfaction by 10%, reduce costs by 15. All within a year of implementing the EHR. \n\nNow look at the options: 

- Potential Solutions: That's what you might do to solve a problem, not the goals. 

- Project Objectives: Objectives are the specific goals a project aims to achieve. That fits because these are the things the project (implementing the EHR) wants to accomplish. 

- Statement of the Problem: That's about the issues or problems that need to be addressed, not the goals. 

- Implementation Plan: That's how you'll carry out the project, like steps or timeline, not the goals themselves. 

So the bullet points are clearly the objectives of the project. The heading should be Project Objectives.
</LONG_COT>
<ANSWER>
Project Objectives
</ANSWER>
\end{lstlisting}

%% file: main.bbl
\begin{thebibliography}{42}
\providecommand{\natexlab}[1]{#1}

\bibitem[{Aggarwal and Welleck(2025)}]{aggarwal2025l1}
Pranjal Aggarwal and Sean Welleck. 2025.
\newblock L1: Controlling how long a reasoning model thinks with reinforcement learning.
\newblock \emph{arXiv preprint arXiv:2503.04697}.

\bibitem[{AIME()}]{aime25}
AIME.
\newblock \href {https://maa.org/math-competitions/american-invitational-mathematics-examination-aime} {Aime problems and solutions}.

\bibitem[{Alomrani et~al.(2025)Alomrani, Zhang, Li, Sun, Pal, Zhang, Hu, Ajwani, Valkanas, Karimi et~al.}]{alomrani2025reasoning}
Mohammad~Ali Alomrani, Yingxue Zhang, Derek Li, Qianyi Sun, Soumyasundar Pal, Zhanguang Zhang, Yaochen Hu, Rohan~Deepak Ajwani, Antonios Valkanas, Raika Karimi, and 1 others. 2025.
\newblock Reasoning on a budget: A survey of adaptive and controllable test-time compute in llms.
\newblock \emph{arXiv preprint arXiv:2507.02076}.

\bibitem[{Bai et~al.(2025)Bai, Chen, Liu, Wang, Ge, Song, Dang, Wang, Wang, Tang et~al.}]{bai2025qwen2}
Shuai Bai, Keqin Chen, Xuejing Liu, Jialin Wang, Wenbin Ge, Sibo Song, Kai Dang, Peng Wang, Shijie Wang, Jun Tang, and 1 others. 2025.
\newblock Qwen2. 5-vl technical report.
\newblock \emph{arXiv preprint arXiv:2502.13923}.

\bibitem[{Chen et~al.(2023)Chen, Ma, Wang, and Cohen}]{chen2022program}
Wenhu Chen, Xueguang Ma, Xinyi Wang, and William~W Cohen. 2023.
\newblock Program of thoughts prompting: Disentangling computation from reasoning for numerical reasoning tasks.
\newblock \emph{Transactions on Machine Learning Research}.

\bibitem[{Chen et~al.(2024)Chen, Xu, Liang, He, Pang, Yu, Song, Liu, Zhou, Zhang et~al.}]{chen2024not}
Xingyu Chen, Jiahao Xu, Tian Liang, Zhiwei He, Jianhui Pang, Dian Yu, Linfeng Song, Qiuzhi Liu, Mengfei Zhou, Zhuosheng Zhang, and 1 others. 2024.
\newblock Do not think that much for 2+ 3=? on the overthinking of o1-like llms.
\newblock \emph{arXiv preprint arXiv:2412.21187}.

\bibitem[{Cobbe et~al.(2021)Cobbe, Kosaraju, Bavarian, Chen, Jun, Kaiser, Plappert, Tworek, Hilton, Nakano et~al.}]{cobbe2021training}
Karl Cobbe, Vineet Kosaraju, Mohammad Bavarian, Mark Chen, Heewoo Jun, Lukasz Kaiser, Matthias Plappert, Jerry Tworek, Jacob Hilton, Reiichiro Nakano, and 1 others. 2021.
\newblock Training verifiers to solve math word problems.
\newblock \emph{arXiv preprint arXiv:2110.14168}.

\bibitem[{Cuadron et~al.(2025)Cuadron, Li, Ma, Wang, Wang, Zhuang, Liu, Schroeder, Xia, Mao et~al.}]{cuadron2025danger}
Alejandro Cuadron, Dacheng Li, Wenjie Ma, Xingyao Wang, Yichuan Wang, Siyuan Zhuang, Shu Liu, Luis~Gaspar Schroeder, Tian Xia, Huanzhi Mao, and 1 others. 2025.
\newblock The danger of overthinking: Examining the reasoning-action dilemma in agentic tasks.
\newblock \emph{arXiv preprint arXiv:2502.08235}.

\bibitem[{Fan et~al.(2025)Fan, Li, Sun, and Zhou}]{fan2025missing}
Chenrui Fan, Ming Li, Lichao Sun, and Tianyi Zhou. 2025.
\newblock \href {https://openreview.net/forum?id=ufozo2Wc9e} {Missing premise exacerbates overthinking: Are reasoning models losing critical thinking skill?}
\newblock In \emph{Second Conference on Language Modeling}.

\bibitem[{Fu et~al.(2024)Fu, Hu, Li, Feng, Wang, Lin, Roth, Smith, Ma, and Krishna}]{fu2024blink}
Xingyu Fu, Yushi Hu, Bangzheng Li, Yu~Feng, Haoyu Wang, Xudong Lin, Dan Roth, Noah~A Smith, Wei-Chiu Ma, and Ranjay Krishna. 2024.
\newblock Blink: Multimodal large language models can see but not perceive.
\newblock In \emph{European Conference on Computer Vision}, pages 148--166. Springer.

\bibitem[{Guo et~al.(2025)Guo, Yang, Zhang, Song, Zhang, Xu, Zhu, Ma, Wang, Bi et~al.}]{guo2025deepseek}
Daya Guo, Dejian Yang, Haowei Zhang, Junxiao Song, Ruoyu Zhang, Runxin Xu, Qihao Zhu, Shirong Ma, Peiyi Wang, Xiao Bi, and 1 others. 2025.
\newblock Deepseek-r1: Incentivizing reasoning capability in llms via reinforcement learning.
\newblock \emph{arXiv preprint arXiv:2501.12948}.

\bibitem[{Hou et~al.(2025)Hou, Zhang, Ji, Liu, Qian, Andreas, and Chang}]{hou2025thinkprune}
Bairu Hou, Yang Zhang, Jiabao Ji, Yujian Liu, Kaizhi Qian, Jacob Andreas, and Shiyu Chang. 2025.
\newblock Thinkprune: Pruning long chain-of-thought of llms via reinforcement learning.
\newblock \emph{arXiv preprint arXiv:2504.01296}.

\bibitem[{Hu et~al.(2022)Hu, Wallis, Allen-Zhu, Li, Wang, Wang, Chen et~al.}]{hu2022lora}
Edward~J Hu, Phillip Wallis, Zeyuan Allen-Zhu, Yuanzhi Li, Shean Wang, Lu~Wang, Weizhu Chen, and 1 others. 2022.
\newblock Lora: Low-rank adaptation of large language models.
\newblock In \emph{International Conference on Learning Representations}.

\bibitem[{Jaech et~al.(2024)Jaech, Kalai, Lerer, Richardson, El-Kishky, Low, Helyar, Madry, Beutel, Carney et~al.}]{jaech2024openai}
Aaron Jaech, Adam Kalai, Adam Lerer, Adam Richardson, Ahmed El-Kishky, Aiden Low, Alec Helyar, Aleksander Madry, Alex Beutel, Alex Carney, and 1 others. 2024.
\newblock Openai o1 system card.
\newblock \emph{arXiv preprint arXiv:2412.16720}.

\bibitem[{Jia et~al.(2025)Jia, Li, Yue, Li, Nie, Zou, and Chen}]{jia2025visualwebinstruct}
Yiming Jia, Jiachen Li, Xiang Yue, Bo~Li, Ping Nie, Kai Zou, and Wenhu Chen. 2025.
\newblock Visualwebinstruct: Scaling up multimodal instruction data through web search.
\newblock \emph{arXiv preprint arXiv:2503.10582}.

\bibitem[{Lightman et~al.(2023)Lightman, Kosaraju, Burda, Edwards, Baker, Lee, Leike, Schulman, Sutskever, and Cobbe}]{lightman2023let}
Hunter Lightman, Vineet Kosaraju, Yuri Burda, Harrison Edwards, Bowen Baker, Teddy Lee, Jan Leike, John Schulman, Ilya Sutskever, and Karl Cobbe. 2023.
\newblock Let's verify step by step.
\newblock In \emph{The Twelfth International Conference on Learning Representations}.

\bibitem[{Ling et~al.(2017)Ling, Yogatama, Dyer, and Blunsom}]{ling2017program}
Wang Ling, Dani Yogatama, Chris Dyer, and Phil Blunsom. 2017.
\newblock Program induction by rationale generation: Learning to solve and explain algebraic word problems.
\newblock In \emph{Proceedings of the 55th Annual Meeting of the Association for Computational Linguistics (Volume 1: Long Papers)}, pages 158--167.

\bibitem[{Loshchilov and Hutter(2017)}]{loshchilov2017sgdr}
Ilya Loshchilov and Frank Hutter. 2017.
\newblock Sgdr: Stochastic gradient descent with warm restarts.
\newblock In \emph{International Conference on Learning Representations}.

\bibitem[{Lou et~al.(2025)Lou, Sun, Liang, Qu, Shen, Wang, Li, Yang, and Wu}]{lou2025adacot}
Chenwei Lou, Zewei Sun, Xinnian Liang, Meng Qu, Wei Shen, Wenqi Wang, Yuntao Li, Qingping Yang, and Shuangzhi Wu. 2025.
\newblock Adacot: Pareto-optimal adaptive chain-of-thought triggering via reinforcement learning.
\newblock \emph{arXiv preprint arXiv:2505.11896}.

\bibitem[{Lu et~al.(2021)Lu, Gong, Jiang, Qiu, Huang, Liang, and Zhu}]{lu2021inter}
Pan Lu, Ran Gong, Shibiao Jiang, Liang Qiu, Siyuan Huang, Xiaodan Liang, and Song-chun Zhu. 2021.
\newblock Inter-gps: Interpretable geometry problem solving with formal language and symbolic reasoning.
\newblock In \emph{Proceedings of the 59th Annual Meeting of the Association for Computational Linguistics and the 11th International Joint Conference on Natural Language Processing (Volume 1: Long Papers)}, pages 6774--6786.

\bibitem[{Luo et~al.(2025{\natexlab{a}})Luo, He, Wang, Yang, Liu, Tan, Cao, Tao, and Shen}]{luo2025ada}
Haotian Luo, Haiying He, Yibo Wang, Jinluan Yang, Rui Liu, Naiqiang Tan, Xiaochun Cao, Dacheng Tao, and Li~Shen. 2025{\natexlab{a}}.
\newblock Ada-r1: Hybrid-cot via bi-level adaptive reasoning optimization.
\newblock \emph{arXiv preprint arXiv:2504.21659}.

\bibitem[{Luo et~al.(2025{\natexlab{b}})Luo, Shen, He, Wang, Liu, Li, Tan, Cao, and Tao}]{luo2025o1}
Haotian Luo, Li~Shen, Haiying He, Yibo Wang, Shiwei Liu, Wei Li, Naiqiang Tan, Xiaochun Cao, and Dacheng Tao. 2025{\natexlab{b}}.
\newblock O1-pruner: Length-harmonizing fine-tuning for o1-like reasoning pruning.
\newblock \emph{arXiv preprint arXiv:2501.12570}.

\bibitem[{Masry et~al.(2022)Masry, Do, Tan, Joty, and Hoque}]{masry2022chartqa}
Ahmed Masry, Xuan~Long Do, Jia~Qing Tan, Shafiq Joty, and Enamul Hoque. 2022.
\newblock Chartqa: A benchmark for question answering about charts with visual and logical reasoning.
\newblock In \emph{Findings of the Association for Computational Linguistics: ACL 2022}, pages 2263--2279.

\bibitem[{Meng et~al.(2025)Meng, Du, Liu, Zhou, Lu, Fu, Shi, Wang, He, Zhang et~al.}]{meng2025mm}
Fanqing Meng, Lingxiao Du, Zongkai Liu, Zhixiang Zhou, Quanfeng Lu, Daocheng Fu, Botian Shi, Wenhai Wang, Junjun He, Kaipeng Zhang, and 1 others. 2025.
\newblock Mm-eureka: Exploring visual aha moment with rule-based large-scale reinforcement learning.
\newblock \emph{arXiv preprint arXiv:2503.07365}.

\bibitem[{Mihaylov et~al.(2018)Mihaylov, Clark, Khot, and Sabharwal}]{mihaylov2018can}
Todor Mihaylov, Peter Clark, Tushar Khot, and Ashish Sabharwal. 2018.
\newblock Can a suit of armor conduct electricity? a new dataset for open book question answering.
\newblock In \emph{Proceedings of the 2018 Conference on Empirical Methods in Natural Language Processing}, pages 2381--2391.

\bibitem[{OpenAI(2024)}]{gpt4o}
OpenAI. 2024.
\newblock \href {https://openai.com/index/hello-gpt-4o/} {Hello {GPT-4o}}.

\bibitem[{Rein et~al.(2024)Rein, Hou, Stickland, Petty, Pang, Dirani, Michael, and Bowman}]{rein2024gpqa}
David Rein, Betty~Li Hou, Asa~Cooper Stickland, Jackson Petty, Richard~Yuanzhe Pang, Julien Dirani, Julian Michael, and Samuel~R Bowman. 2024.
\newblock Gpqa: A graduate-level google-proof q\&a benchmark.
\newblock In \emph{First Conference on Language Modeling}.

\bibitem[{Seed et~al.(2025)Seed, Chen, Fan, Liu, Liu, Lin, Wang, Wang, Wei, Xu et~al.}]{seed2025seed1}
ByteDance Seed, Jiaze Chen, Tiantian Fan, Xin Liu, Lingjun Liu, Zhiqi Lin, Mingxuan Wang, Chengyi Wang, Xiangpeng Wei, Wenyuan Xu, and 1 others. 2025.
\newblock Seed1. 5-thinking: Advancing superb reasoning models with reinforcement learning.
\newblock \emph{arXiv preprint arXiv:2504.13914}.

\bibitem[{Shao et~al.(2024)Shao, Wang, Zhu, Xu, Song, Bi, Zhang, Zhang, Li et~al.}]{shao2024deepseekmath}
Zhihong Shao, Peiyi Wang, Qihao Zhu, Runxin Xu, Junxiao Song, Xiao Bi, Haowei Zhang, Mingchuan Zhang, YK~Li, and 1 others. 2024.
\newblock Deepseekmath: Pushing the limits of mathematical reasoning in open language models.
\newblock \emph{arXiv preprint arXiv:2402.03300}.

\bibitem[{Sheng et~al.(2024)Sheng, Zhang, Ye, Wu, Zhang, Zhang, Peng, Lin, and Wu}]{sheng2024hybridflow}
Guangming Sheng, Chi Zhang, Zilingfeng Ye, Xibin Wu, Wang Zhang, Ru~Zhang, Yanghua Peng, Haibin Lin, and Chuan Wu. 2024.
\newblock Hybridflow: A flexible and efficient rlhf framework.
\newblock \emph{arXiv preprint arXiv: 2409.19256}.

\bibitem[{Sui et~al.(2025)Sui, Chuang, Wang, Zhang, Zhang, Yuan, Liu, Wen, Zhong, Chen et~al.}]{sui2025stop}
Yang Sui, Yu-Neng Chuang, Guanchu Wang, Jiamu Zhang, Tianyi Zhang, Jiayi Yuan, Hongyi Liu, Andrew Wen, Shaochen Zhong, Hanjie Chen, and 1 others. 2025.
\newblock Stop overthinking: A survey on efficient reasoning for large language models.
\newblock \emph{arXiv preprint arXiv:2503.16419}.

\bibitem[{Talmor et~al.(2019)Talmor, Herzig, Lourie, and Berant}]{talmor2019commonsenseqa}
Alon Talmor, Jonathan Herzig, Nicholas Lourie, and Jonathan Berant. 2019.
\newblock Commonsenseqa: A question answering challenge targeting commonsense knowledge.
\newblock In \emph{Proceedings of the 2019 Conference of the North American Chapter of the Association for Computational Linguistics: Human Language Technologies, Volume 1 (Long and Short Papers)}, pages 4149--4158.

\bibitem[{Wang et~al.(2024)Wang, Chen, Yuan, Zhang, Li, Peng, and Ji}]{wang2024executable}
Xingyao Wang, Yangyi Chen, Lifan Yuan, Yizhe Zhang, Yunzhu Li, Hao Peng, and Heng Ji. 2024.
\newblock Executable code actions elicit better llm agents.
\newblock In \emph{Forty-first International Conference on Machine Learning}.

\bibitem[{Wang et~al.(2023)Wang, Wei, Schuurmans, Le, Chi, Narang, Chowdhery, and Zhou}]{wang2023self}
Xuezhi Wang, Jason Wei, Dale Schuurmans, Quoc~V Le, Ed~H Chi, Sharan Narang, Aakanksha Chowdhery, and Denny Zhou. 2023.
\newblock Self-consistency improves chain of thought reasoning in language models.
\newblock In \emph{The Eleventh International Conference on Learning Representations}.

\bibitem[{Wu et~al.(2024)Wu, Xie, Chen, Zhu, Zhang, and Xiao}]{wu2024how}
Siye Wu, Jian Xie, Jiangjie Chen, Tinghui Zhu, Kai Zhang, and Yanghua Xiao. 2024.
\newblock How easily do irrelevant inputs skew the responses of large language models?
\newblock In \emph{First Conference on Language Modeling}.

\bibitem[{Wu et~al.(2025)Wu, Xie, Zhang, Chen, Zhang, Su, and Xiao}]{wu2025arm}
Siye Wu, Jian Xie, Yikai Zhang, Aili Chen, Kai Zhang, Yu~Su, and Yanghua Xiao. 2025.
\newblock Arm: Adaptive reasoning model.
\newblock \emph{arXiv preprint arXiv:2505.20258}.

\bibitem[{Xiaomi(2025)}]{coreteam2025mimovltechnicalreport}
LLM-Core-Team Xiaomi. 2025.
\newblock \href {https://arxiv.org/abs/2506.03569} {Mimo-vl technical report}.
\newblock \emph{Preprint}, arXiv:2506.03569.

\bibitem[{Yue et~al.(2025)Yue, Du, Wang, Gao, Yao, Wang, Liu, Xu, Liu, Di et~al.}]{yue2025don}
Linan Yue, Yichao Du, Yizhi Wang, Weibo Gao, Fangzhou Yao, Li~Wang, Ye~Liu, Ziyu Xu, Qi~Liu, Shimin Di, and 1 others. 2025.
\newblock Don't overthink it: A survey of efficient r1-style large reasoning models.
\newblock \emph{arXiv preprint arXiv:2508.02120}.

\bibitem[{Yue et~al.(2024)Yue, Ni, Zhang, Zheng, Liu, Zhang, Stevens, Jiang, Ren, Sun et~al.}]{yue2024mmmu}
Xiang Yue, Yuansheng Ni, Kai Zhang, Tianyu Zheng, Ruoqi Liu, Ge~Zhang, Samuel Stevens, Dongfu Jiang, Weiming Ren, Yuxuan Sun, and 1 others. 2024.
\newblock Mmmu: A massive multi-discipline multimodal understanding and reasoning benchmark for expert agi.
\newblock In \emph{Proceedings of the IEEE/CVF Conference on Computer Vision and Pattern Recognition}, pages 9556--9567.

\bibitem[{Zhang et~al.(2025)Zhang, Lin, Hou, Feng, and Li}]{zhang2025adaptthink}
Jiajie Zhang, Nianyi Lin, Lei Hou, Ling Feng, and Juanzi Li. 2025.
\newblock Adaptthink: Reasoning models can learn when to think.
\newblock \emph{arXiv preprint arXiv:2505.13417}.

\bibitem[{Zhang et~al.(2024)Zhang, Zhang, Tian, Fu, Zhang, Wu, Li, Wang, Wen, Zhang et~al.}]{zhang2024mme}
YiFan Zhang, Huanyu Zhang, Haochen Tian, Chaoyou Fu, Shuangqing Zhang, Junfei Wu, Feng Li, Kun Wang, Qingsong Wen, Zhang Zhang, and 1 others. 2024.
\newblock Mme-realworld: Could your multimodal llm challenge high-resolution real-world scenarios that are difficult for humans?
\newblock In \emph{The Thirteenth International Conference on Learning Representations}.

\bibitem[{Zhu and Li(2025)}]{zhu2025towards}
Jason Zhu and Hongyu Li. 2025.
\newblock Towards concise and adaptive thinking in large reasoning models: A survey.
\newblock \emph{arXiv preprint arXiv:2507.09662}.

\end{thebibliography}
